%% file: main.tex
\documentclass[lettersize,journal]{IEEEtran}
\IEEEoverridecommandlockouts
\usepackage{cite}
\usepackage{amsmath,amssymb,amsfonts}
\usepackage{algorithm}
\usepackage{algpseudocode}
\usepackage{graphicx}
\usepackage{textcomp}
\usepackage{mdframed}
\usepackage[table,dvipsnames]{xcolor}
\usepackage{multirow}
\usepackage[hyphens]{url}
\usepackage{hyperref}
\usepackage{tikz}
\usepackage{subfig,xspace}
\usepackage{caption}
\usepackage[most]{tcolorbox}

\usepackage[skip=1pt]{caption}
\newcommand*\circled[1]{\tikz[baseline=(char.base)]{\node[shape=circle,fill,inner sep=0pt,minimum size=1pt] (char) {\textcolor{white}{#1}};}}
\usepackage[a4paper, total={184mm,239mm}]{geometry}
\def\BibTeX{{\rm B\kern-.05em{\sc i\kern-.025em b}\kern-.08em
    T\kern-.1667em\lower.7ex\hbox{E}\kern-.125emX}}

\newcommand{\voxdepth}{{\em VoxDepth}\xspace}
\begin{document}

\author{
        Yashashwee Chakrabarty
        \and
        Smruti R Sarangi\\

        \textit{yashashwee99@gmail.com}
        \and
        \textit{srsarangi@cse.iitd.ac.in}
}

\title{\voxdepth: Rectification of Depth Images on Edge Devices}

\maketitle

\begin{abstract} 
Autonomous mobile robots like self-flying drones and industrial robots heavily depend on depth images
to perform tasks such as 3D reconstruction and visual SLAM. 
However, the presence of inaccuracies in these depth images
can greatly hinder the effectiveness of these applications, resulting in sub-optimal results.
Depth images
produced by commercially available cameras frequently exhibit noise, which manifests as flickering pixels and erroneous
patches.  ML-based methods to rectify these images are unsuitable for
edge devices that have very limited computational resources. Non-ML methods are much
faster but have limited accuracy, especially for correcting errors that are a result of
occlusion and camera movement.
We propose a scheme called \voxdepth that is fast, accurate, and runs very well on 
edge devices. It relies on a host of novel techniques: 3D point cloud construction
and fusion, and using it to create a template that can fix erroneous depth images.
\voxdepth shows superior results on both synthetic and
real-world datasets. 
We demonstrate
a $31\%$ improvement in quality as compared to state-of-the-art methods on real-world depth datasets,
while maintaining a competitive framerate of $27$ FPS (frames per second).\\
\end{abstract}


\input{introduction}
\input{background}
\input{characterization}
\input{implementation}

\input{evaluation}
\input{relatedwork}
\input{conclusion}
\newpage

\bibliographystyle{IEEEtranS}
\bibliography{references}

\end{document}

%% file: introduction.tex
\section{Introduction}
\label{sec:Introduction}
The\footnote{Kindly take a color printout or use a color monitor to view the PDF file. A monochrome printout will not be able to show the key features of images and our contributions.} estimated market size of autonomous mobile robots is USD 3.88 billion in 2024, which is projected to reach USD 8.02 billion by 2029. 
This growth is expected to occur at a compound annual growth rate (CAGR) of 
$15.60\%$ throughout~\cite{amrm}. 
For such autonomous systems,
high-quality depth images are crucial. They are needed to support different
robotics applications including RGB-D SLAM~\cite{dai2020rgb}
(simultaneous localization and mapping), 3D object detection~\cite{song2014sliding}, drone
swarming~\cite{bhamu2023smrtswarm}, 3D environment mapping and surveillance. 
Depth images are produced by depth
cameras, which are specialized cameras explicitly
designed for this purpose.
Depth images are a matrix of integer value, each representing the depth of the corresponding pixel in the color image.
Various types of depth cameras exist, each based on a
different working principle.  Three of the most popular types are shown in Table~\ref{tab:comparestereo}.

\begin{table}[!htb]
  \centering
  \caption{Comparison of different depth imaging acquisition methods}
  \label{tab:depth_devices_comparison}
  \resizebox{1.0\columnwidth}{!}{
  \begin{tabular}{|l|l|l|l|}
  \hline
  \textbf{Aspect} & \textbf{Stereo Cameras} & \textbf{LiDAR Sensors} & \textbf{Structured Light} \\ 
  \hline
  Accurate depth perception & \textcolor{ForestGreen}{High} & \textcolor{ForestGreen}{High} & \textcolor{ForestGreen}{High} \\ 
  \hline
  Range & \textcolor{orange}{Medium} & \textcolor{ForestGreen}{High} & \textcolor{red}{Low} \\ 
  \hline
  Resolution & \textcolor{ForestGreen}{High} & \textcolor{red}{Low} & \textcolor{orange}{Medium} \\ 
  \hline
  Environmental adaptability & \textcolor{orange}{Medium} & \textcolor{ForestGreen}{High} & \textcolor{red}{Low} \\ 
  \hline
  Affordability & \textcolor{ForestGreen}{High} & \textcolor{red}{Low} & \textcolor{orange}{Medium} \\ 
  \hline
  Minimal energy consumption & \textcolor{ForestGreen}{High} & \textcolor{red}{Low} & \textcolor{orange}{Medium} \\ 
  \hline
  Light weight and compact & \textcolor{ForestGreen}{High} & \textcolor{red}{Low} & \textcolor{orange}{Medium} \\ 
  \hline
  \end{tabular}
  }
  \label{tab:comparestereo}
\end{table}
\begin{figure}[!htb]
\subfloat[Color]{
\includegraphics[width=0.50\columnwidth]{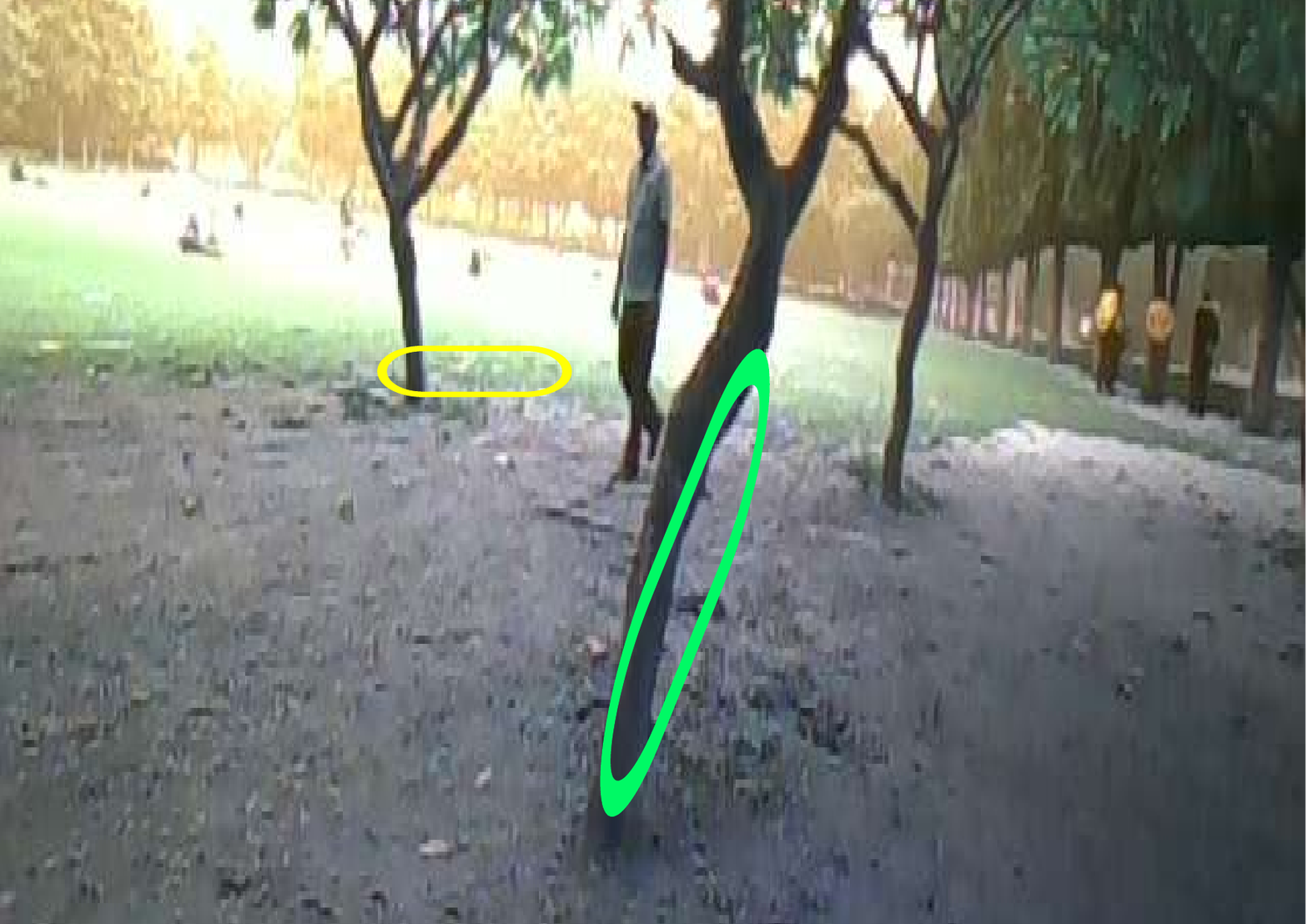}
}
\subfloat[Corrected]{
\includegraphics[width=0.50\columnwidth]{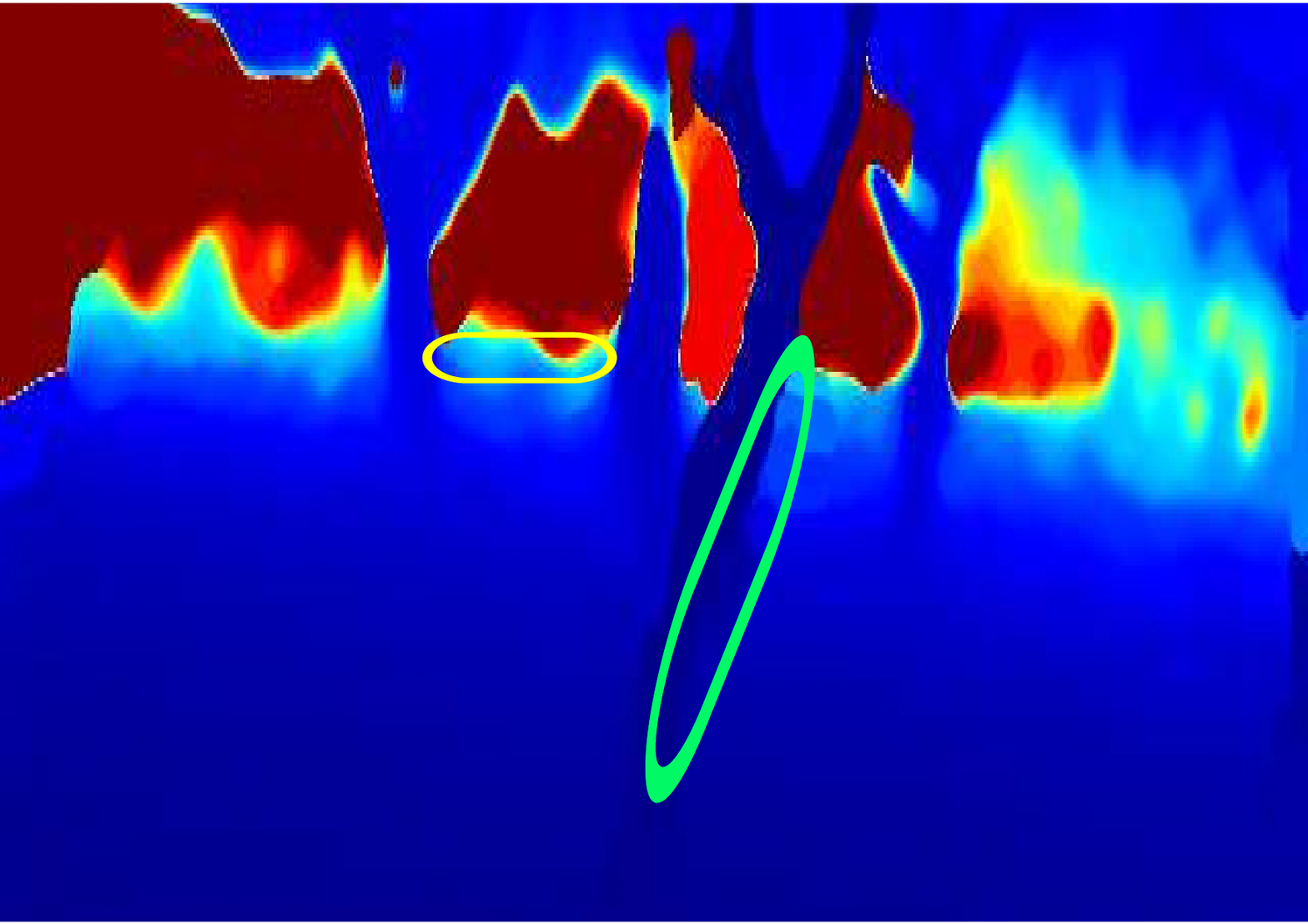}
}
\vspace{-4mm}
\subfloat[Frame 0]{
\includegraphics[width=0.50\columnwidth]{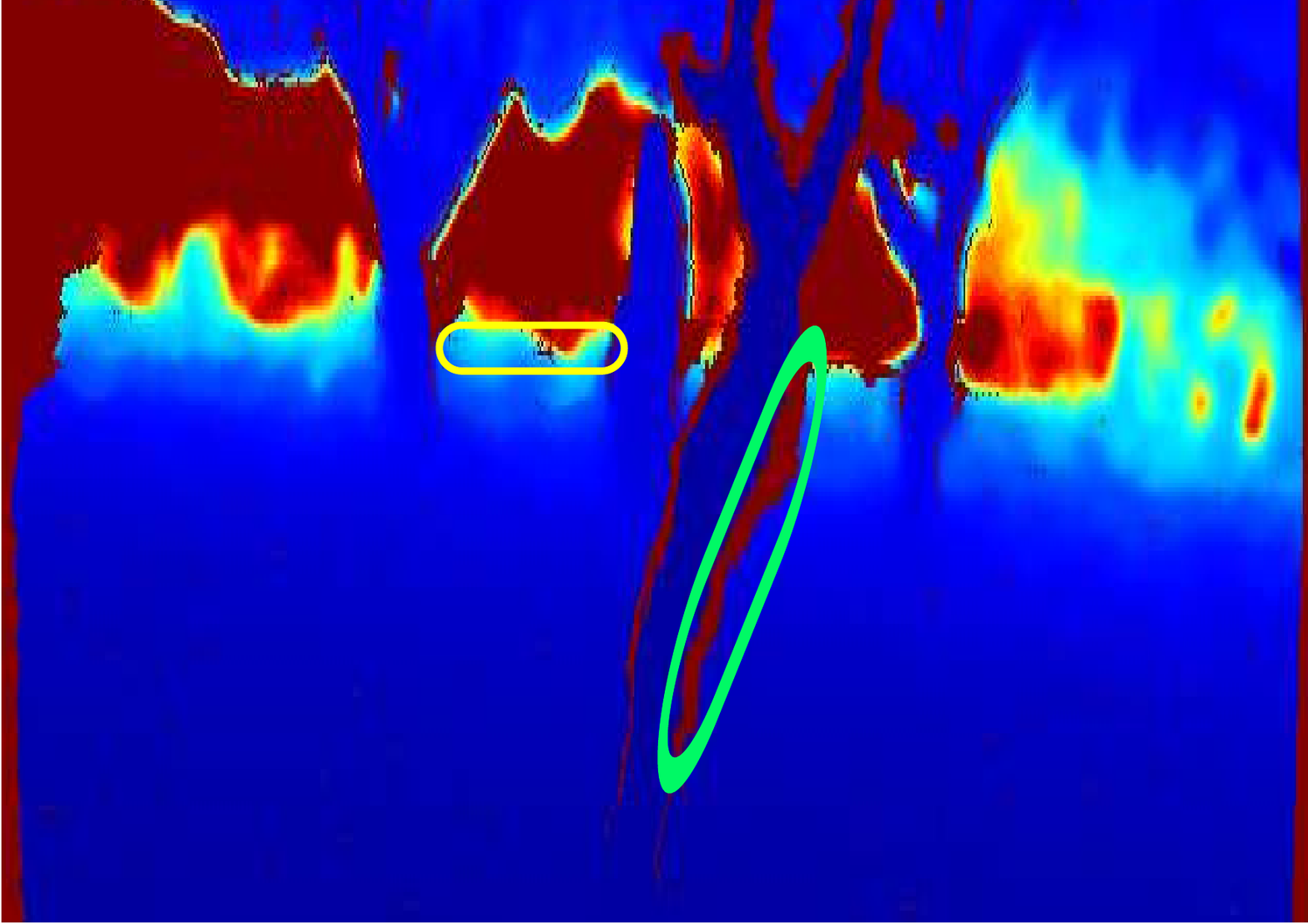}
}
\subfloat[Frame 1]{
\includegraphics[width=0.50\columnwidth]{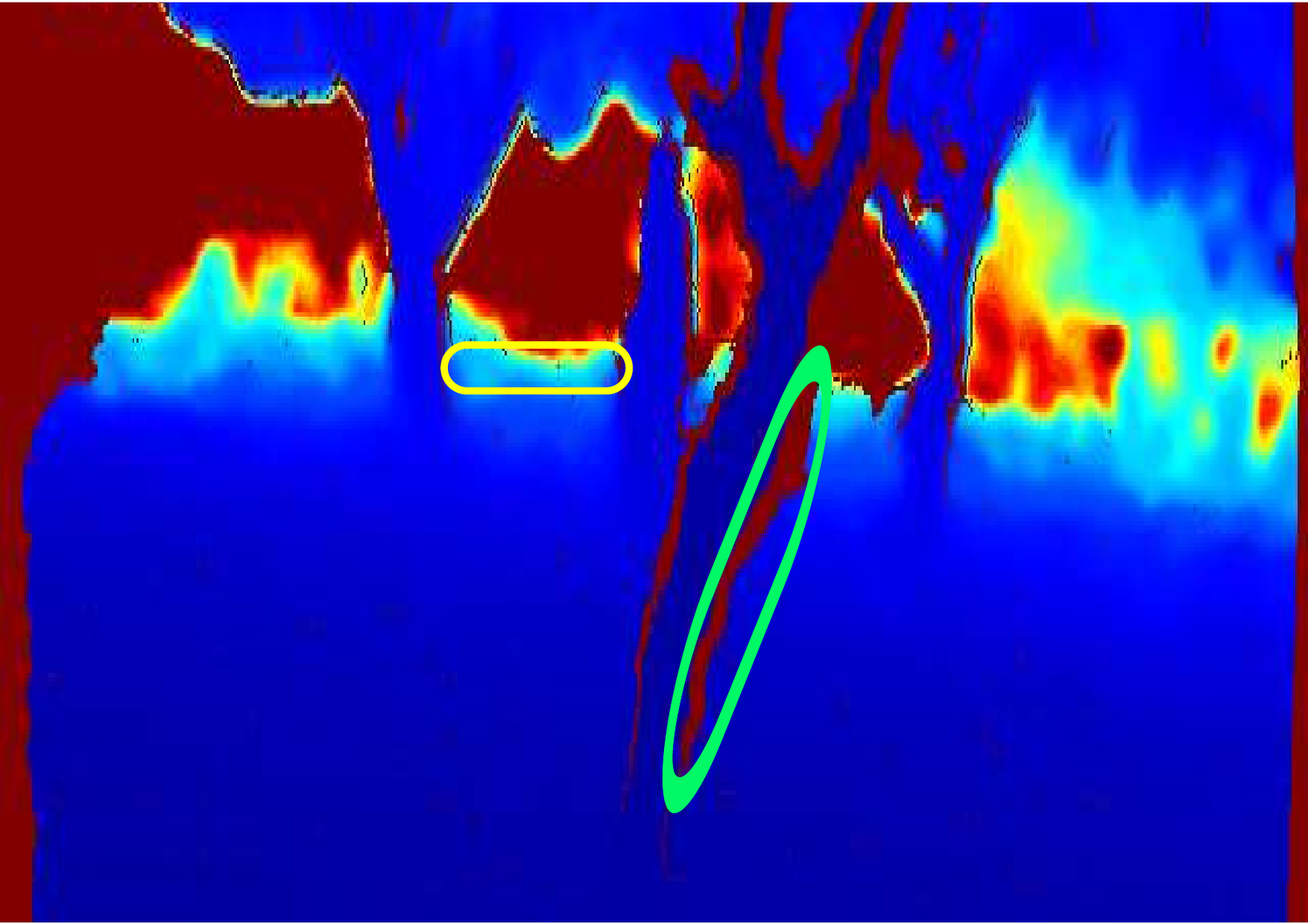}
}
\caption{Visual representation of algorithmic holes (green ellipse) and flickering noise (yellow ellipse) across two frames. The flickering noise appears and disappears across frames, but algorithmic noise persists across frames as long as the object remains present.}
\label{noisedesc}
\end{figure}

Mobile robot developers have increasingly favored stereoscopic depth cameras (or just stereo cameras) due to their
compact size, reduced weight and the reasons mentioned in Table~\ref{tab:depth_devices_comparison}.  These depth cameras
are equipped with specialized electronics that can provide a continuous flow of high-resolution depth images at a rate
of $90$ RGB-D (pairs of aligned color and depth images) frames per second. Other methods of depth acquisition,
such as those listed in Table \ref{tab:comparestereo}, are more challenging to implement on lightweight autonomous
systems like drones because they require more weight and power. Economically as well, stereo cameras are a better choice
for small-scale mobile robotic systems.  The popular small-scale LiDAR sensor Velodyne Puck LITE costs around $\$ 5000 $
USD, while the similarly sized Intel RealSense D455 camera is listed for $\$409 $ USD on Amazon and is also $200$grams
lighter.  There are several popular stereoscopic depth cameras available in the market today, such as the ZED 2 depth
camera, which has a price similar to that of the D455, but most of them lack the ability to generate frames at a high framerate and produce
high-resolution depth images as compared to the Intel offering. 

Sadly, the depth images generated by such depth cameras are prone to imperfections such as noise and gaps.
Inaccuracies in depth images result in unfavorable consequences, particularly in tasks involving close
contact~\cite{lasota2014toward}.  An erroneous perception of depth has the potential to result in collisions, thereby
causing damage to property and perhaps endangering humans. Multiple news
articles~\cite{Thadani_Siddiqui_Lerman_Merrill_2023,Uraizee_2023} have  documented such incidents.  In Section
\ref{subsec:consequences}, we shall show how sensitive drone swarming is to errors in depth data. We shall observe that
the collision rate increases super-linearly with the error.

The process of estimating stereoscopic depth is hindered by two primary forms of inaccuracy in the depth images 
(refer to Figure~\ref{noisedesc}): 
\circled{1} flickering pixel noise and \circled{2} algorithmic holes
that arise from failures in stereoscopic matching.  
The first type of noise is random in character~\cite{ibrahim2020depth} and may be corrected using
classical spatial filters, whereas the second type requires far more complex solutions. We
shall define algorithmic holes in more detail 
in Section \ref{subsubsec:algonoise}.

\subsection*{\bf Shortcomings in Prior Work}
Prior research in this domain can be broadly classified into two distinct categories: \circled{1} ML-based methods ~\cite{krishna2023deepsmooth,zhang2018deep,xu2019depth} and \circled{2} non-ML methods ~\cite{islam2018gsmooth,grunnet2018depth,avetisyan2016temporal,matyunin2011temporal}. 
There are two problems with ML-based methods, notably neural networks. The availability of a sufficient 
amount of real-world data is an issue. There are many proposed methods that expand the size of datasets by 
creating new images based on the images that have already been collected~\cite{sterzentsenko2019self,senushkin2021decoder}.
However, their applicability and generalizability is limited.
The other major issue is that
even the most optimized networks in this space do not attain a frame rate that is more than $2-3$ frames-per-second,
which is 10X lower than what is needed.

Non-machine learning methods, in contrast, require only a minimal amount of precise ground truth data for calibration.
The majority of non-machine learning techniques employ spatial ~\cite{islam2018gsmooth} and temporal
~\cite{avetisyan2016temporal} filters on frames to eliminate noisy pixels. These approaches provide a high level of
efficiency and accuracy when dealing with flickering noise. However, the image quality is quite low when it 
comes to
algorithmic noise (holes).

Given that the state-of-the-art solutions either lack in terms of speed (ML) or quality (non-ML), we can conclude that
there is a need for a new depth image rectification technique.  It needs to respect many more practical
constraints such as 
a limit on the cost of the sensor and computing board, limit on the maximum
power usage and the weight of the assemblage.  Our chosen
setting is an NVIDIA Jetson Nano board that is quite popular in this 
space~\cite{assunccao2022real,kim2021study}. It is
meant to be mounted on a 2-3 kg drone or mobile robot.

\subsection*{\bf Constraints and Objective Function}
Our objective function is to achieve a frame rate of at least 20 FPS (deemed to be sufficient
in our simulations and prior work~\cite{delmerico2019we}) on a Jetson Nano board and improve the depth estimation quality vis-à-vis the state of the art.
The constraints are as follows: the average power usage must not exceed 5 Watts -- \textit{mode $1$} of the Jetson Nano board~\cite{TegraLinuxDriver}. The total cost of the camera and the board needs to be in the ballpark of
$\$500-\$700$ USD (2024 pricing), and the added weight must not exceed 700-750 grams.

\subsection*{\bf Contributions}
Here is our technique \voxdepth in a nutshell. We propose a novel 3D representation of the scene that is stored as a 
{\em point cloud}. This point cloud is repeatedly updated as new scenes arrive. We have several novel solutions in this space in terms of the representation of the point cloud itself and the methods to update it. Once this is done at the beginning of an epoch, the point cloud
is expected to remain {\em stable} till the end of the epoch. Then we create a 2D template scene out of this ``fused''
point cloud. Whenever a new 2D image arrives, we compare it with the 2D template and identify the regions that may have
suffered from any kind of noise. Algorithmic holes are corrected using information derived from the template, whereas
flickering noise is corrected using standard filters. When the ambience sufficiently changes, a new epoch begins, and
the fused point cloud is computed yet again. This novel method allows us to sustain a frame rate of 27 FPS (at least
70\% more than prior work) and outperform the state-of-the-art in terms of depth estimation quality. 
 
The novel aspects in the design of \voxdepth are as follows:
\circled{1} A point cloud fusion method that aims to combine depth information from a set of consecutive RGB-D frames into a single sparse point cloud.\\
\circled{2} A depth image inpainting method that is used to create a high-resolution depth image from a low-resolution point cloud; it is meant to be used as a 2D scene template.\\
\circled{3} A pipelined module for combining the foreground and background to produce precise 2D depth images from a depth image. The process involves \circled{a} resizing the images, \circled{b} estimating motion, \circled{c} correcting the incoming frame with the 2D scene template in order to produce depth images that exhibit a high level of accuracy.\\
\circled{4} A technique to dynamically recompute the point cloud when the scene changes to a sufficient extent.\\

The code for the project is available at the following git repo: \url{https://github.com/srsarangi/voxdepthcode}.

\subsection*{\bf Outline}
\label{subsec:outline}
We begin with a detailed explanation of stereoscopic depth estimation and its errors (Section~\ref{sec:Background}). Our experimental setting and datasets are described in the
characterization section (Section~\ref{sec:Characterization}). Section~\ref{sec:Implementation} describes our two-step depth image correction approach. To support
our claims, we evaluate our design in Section~\ref{sec:Evaluation}. We present the related work in Section~\ref{sec:RelatedWork}
and finally conclude in Section~\ref{sec:Conclusion}.

%% file: background.tex
\section{Background}
\label{sec:Background}
This section offers the fundamental basis needed to develop and execute the concepts discussed in this paper.
This section consists of two subsections. 
The first subsection provides an explanation on the concept of {\em image matching}, which involves identifying similar regions in two or more images depicting the same scene. This subsection also provides a comprehensive explanation on several significant applications of this approach.
In the subsequent subsection, we introduce a data structure known as a {\em point cloud}, which is a three-dimensional representation of depth images. We also explain the process of merging several point clouds obtained from distinct depth images of the same scene into a unified and densely populated point cloud.
\input{imagematching}
\input{pointcloud}

%% file: imagematching.tex
\subsection{Image Matching: Theory, Applications and Shortcomings }

\subsubsection{Image Matching}
\label{subsubsec:imagematching}
Image matching is a process in computer vision that involves finding {\em correspondences} between two or more images 
of the same scene or object. The goal is to identify points or regions in one image that match corresponding 
points or regions in another image. This is essential for tasks such as image stitching, stereoscopic depth 
estimation, and 3D reconstruction. The methods used to implement such image matching comprise of several 
interlinked steps but these steps can be reduced down to the following three stages:
\begin{enumerate}
    \item \textbf{Feature Detection} involves identifying distinctive keypoints or features in the images. Two widely used methods for feature detection are SURF~\cite{bay2006surf} and FAST~\cite{viswanathan2009features}.
    \item \textbf{Feature Description} creates descriptors for the detected features, which are robust to changes in scale, rotation, and illumination. The most commonly used feature descriptor is the BRIEF~\cite{calonder2010brief} descriptor.
    \item \textbf{Feature Matching} uses the features with their accompanying descriptors to find correspondence. This is a time consuming task and brute force solutions might lead to a sluggish system. Hence approximate matchers such as FLANN~\cite{muja2009flann} are used to speed up the process.
\end{enumerate}
Most matching applications estimate a transformation between the images.
In image processing, a transformation refers to the process of applying mathematical operations to an image to change its characteristics or extract specific information, such as altering its geometry, adjusting its intensity values, or converting it to a different domain.
The matching complexity and type of transformation generated varies according to the use-case. The three applications covered in the following sections include:
\begin{itemize}
    \item \textbf{Translation Transform}: Stereoscopic depth estimation generates a translational transformation in one-dimension between each matching pixel which is used in depth cameras.
    \item \textbf{Rigid Body Transform}: RGB-D odometry estimates the transformation between RGB-D images with six degrees of freedom.
    \item \textbf{Affine Transform}: Image registration estimates a general correspondence between the images using the matching pixels which may allow for shifting, scaling and rotational operations apart from translational.
\end{itemize}

\subsubsection{Translation Transform: Stereoscopic Depth Estimation}
\label{subsubsec:stereodepth}

\begin{figure}[!htb]
    \centering
    \includegraphics[width=0.75\columnwidth]{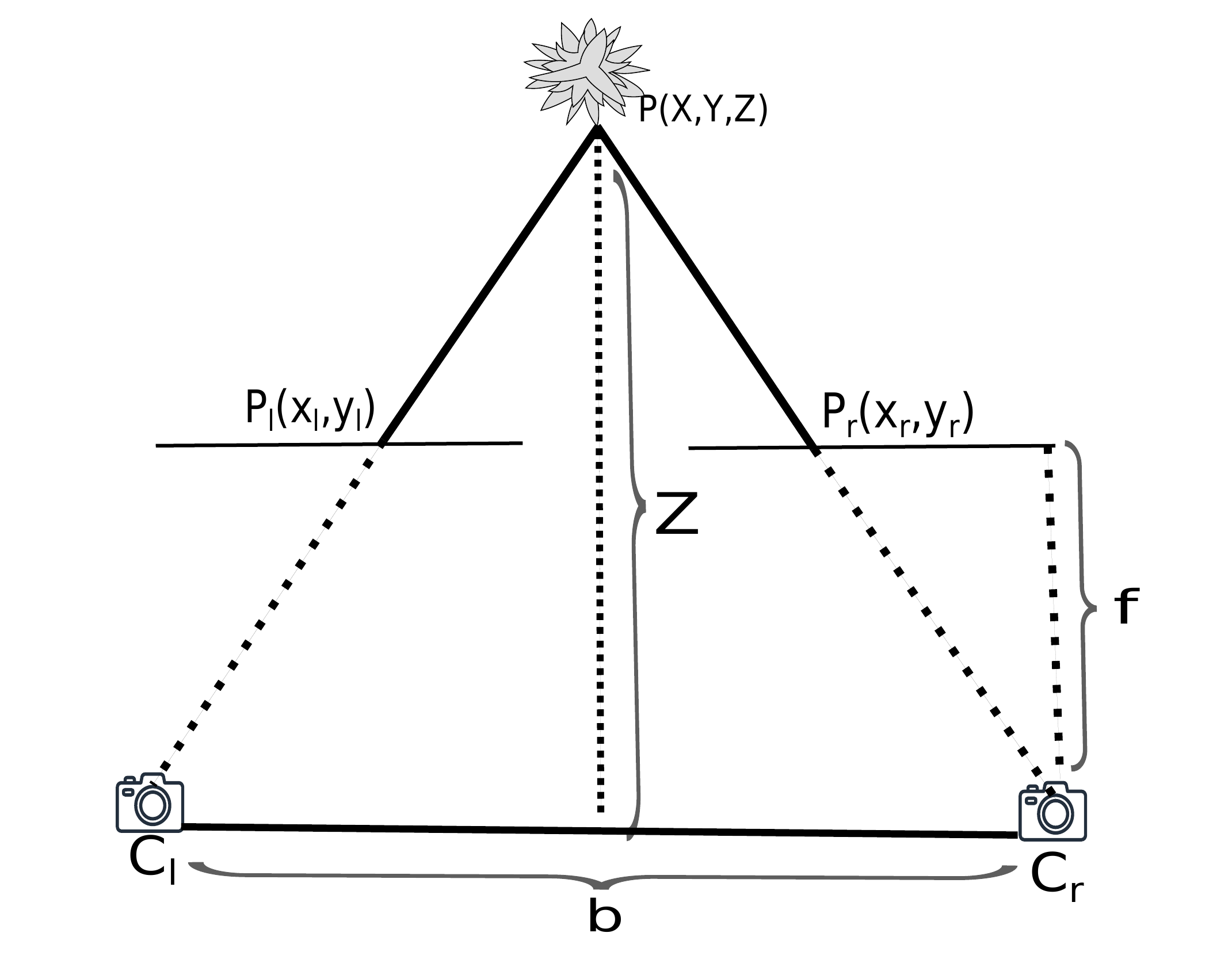}
    \caption{Visual representation of the stereoscopic depth estimation method}
    \label{fig:stereo}
\end{figure}

A Translation transform is a form of mapping between pixels of two images that only allows translational movement of pixels in one dimension. This is useful in stereoscopic depth estimation.
The concept of stereoscopic depth estimation\cite{lane1994stereo} involves employing two distinct sensors (cameras) to observe the same scene from two different angles.
The distance between the two sensors is known and is referred to as the {\em baseline} ($b$).
They are positioned on the same plane and  
synchronized to capture a scene simultaneously. 
Each pixel in the frame captured by the first sensor, $C_l$, is then matched with 
its counterpart in the frame captured by the other sensor, $C_r$. For a point in the scene, say $P$ with coordinates $(x,y,z)$ in the camera's frame of reference, 
each sensor would see a projection in its respective image plane. The distance (in pixels) between a point $P_l$ in the image plane of sensor $C_l$ and 
its corresponding point in the image plane of $C_r$ ($P_r$) is referred to as
the {\em disparity} (refer to Figure \ref{fig:stereo}). 
This is the transform that maps each pixel in one image to each pixel in the other. 
We only consider the $x$ coordinates because the
$y$ coordinates are the same for parallel sensors. Using the property of similar triangles --
triangles $\Delta PP_lP_r$ and $\Delta PC_lC_r$ --
we arrive at the following relation:

\begin{equation}
    \label{eqn:depthest}
        \dfrac{b}{z} = \dfrac{b+x_l-x_r}{z-f}
\end{equation}
By rearranging we get:
\begin{equation}
    \label{eqn:depthest}
        z = \dfrac{f\times b}{(x_r-x_l)}
\end{equation}

Here $z$ is the depth of the point $P$, $f$ is the cameras' focal length and $b$ is the baseline. These depth
images contain substantial noise in the form of \em algorithmic \em and \em flickering \em noise.

\subsubsection{Rigid Body Transform: RGB-D Odometry}
\label{subsubsec:RGBDodo}
The objective of RGB-D 
odometry is to estimate the rigid body motion $q \in SE(3)$ of the camera given two consecutive images $I(t_0)$ and $I(t_1)$.
This {\em motion} is a transformation with six degrees of freedom: three describe camera rotation (roll, pitch and yaw) and
three represent translation ($x,y,z$).  
Using this transformation, we can align multiple point clouds to a common coordinate system, we
can fuse them together to form a {\em dense point cloud}.

\subsubsection{Affine Transform: Image Registration}
\label{subsubsec:imgreg}
An affine transformation refers to a
geometric transformation that preserves points, straight
lines and planes.  It includes operations such as translation 
(shifting), rotation, scaling (resizing) and shearing
(stretching). It is a common transformation used in image 
registration.
Image {\em registration} refers to the process of \underline{aligning}
two distinct images or frames of a shared scene to a common
coordinate system.  In order to accomplish this, we calculate 
the affine transformation between two frames by comparing the
pixels in the two frames.

An affine transformation can be represented as a
rotation $A$ followed by a translation $b$.  The transformed 
coordinates of the image are as follows:

\begin{equation}
    \label{eqn:trans}
        P_1= A\times P_0 + b
\end{equation}

$P_0$ represents the original coordinates of the pixel, 
and $P_1$ represents the computed (transformed) coordinates. 
In our implementation, 
we condense this transformation using a single $2\times3$ matrix $M$.

To estimate an affine transformation, the squared difference between the real image and 
the transformed image is minimized.
Formally, if an image is represented as $I(u,v)$, denoting the pixel intensity at the pixel coordinates $(u,v)$,
then we minimize the following objective function: 

\begin{equation}
    \label{eqn:objfun}
    E = \sum_{u,v} \left\|I_1\left(M \begin{bmatrix} u \\ v \\ 1 \end{bmatrix}\right) - I_0\left(\begin{bmatrix} u \\ v \end{bmatrix}\right)\right\|^2
\end{equation}
Image registration aids in estimating the motion of the camera and transforming frames compensates for that motion to bring frames taken at different 
points of time to a common coordinate system.

\subsubsection{Shortcomings: Algorithmic Noise}
\label{subsubsec:algonoise}
\begin{figure}[!htb]
\centering
  \subfloat[GT]{
    \includegraphics[width=0.40\columnwidth]{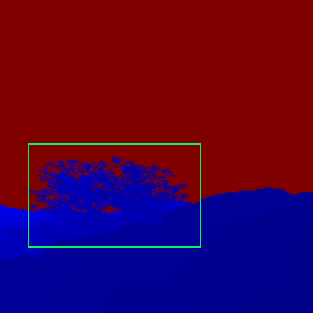}
  }
  \subfloat[With Noise]{
    \includegraphics[width=0.40\columnwidth]{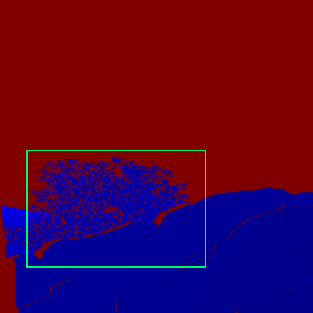}
  }
  \caption{Visual representation of algorithmic noise in an image from the
  Mid-air dataset~\cite{fonder2019mid}. It is shown using a green box.}
  \label{fig:algonoise}
  \end{figure}

{\em Algorithmic noise} refers to the presence of inaccurate regions in depth images resulting from the inability to locate corresponding pixels captured by the other camera.
This can happen due to a couple of reasons: \circled{1} occlusion of the point in one of the cameras; or \circled{2}
lack of textural information leading to patches of 
invalid pixels, which appear as red shadows in the depth images in
Figure~\ref{fig:algonoise} (refer to the green box).

\subsubsection{Shortcomings: Flickering Noise}
\label{subsubsec:flickernoise} 
Flickering noise in depth images refers to rapid and irregular variations in pixel
intensity, often appearing as random fluctuations or 
shimmering effects in the image. This type of noise is commonly
caused by sensor imperfections such as fluctuations
in the sensitivity or exposure time, leading to inconsistent depth
readings and pixel values. These noise artifacts can be corrected by local filters.  They can also happen due to camera
motion.

%% file: pointcloud.tex
\subsection{3D Point Cloud \& Fusion}
\subsubsection{Point Cloud}
\label{subsubsec:pcd}
Point clouds are three-dimensional data structures composed of points representing the surface of objects or scenes in
the real world.
Let us mathematically represent the RGB-D frames generated by a stereoscopic depth camera.
$I_{RGB}$ represents
the channel-wise RGB intensity and $I_{D}$ is the 
depth of the point in meters 
in the image plane $\gamma \subseteq \mathbb{R}_{+}^{2}$ at point $x\in \gamma$, at time $t \in \mathbb{R}_{+}$.

\begin{equation}
    \label{eqn:rgbframe}
    I_{RGB}:\gamma \times \mathbb{R}_{+} \rightarrow [0,1]^{3}, (x,t) \rightarrow I_{RGB}(x,t)
\end{equation}
\begin{equation}
    \label{eqn:dframe}
    I_{D}:\gamma \times \mathbb{R}_{+} \rightarrow \mathbb{R}_{+}, (x,t) \rightarrow I_{D}(x,t)
\end{equation}

From the depth image $I_D$, we can compute the surface $S$ visible from the sensor by projecting the
points in a 3D space for each point $p=(p_x,p_y)\in\gamma$, $\gamma \subseteq \mathbb{R}_{+}^{2}$ in the depth image at
some point at time $t_0$ (see Equation~\ref{eqn:surface2}). In Equation~\ref{eqn:surface2}, $(o_u,o_v)^{T}$ is the
principal point of the camera on the $x$ and $y$ axis, respectively.

\begin{equation}
    \footnotesize
    \label{eqn:dframe}
    I_{D}:\gamma \times \mathbb{R}_{+} \rightarrow \mathbb{R}_{+}, (p,t) \rightarrow I_{D}(p,t)
\end{equation} 
\begin{equation}
    \footnotesize
    \label{eqn:surface1}
    S:\gamma \rightarrow R^{3},
\end{equation}
\begin{equation}
    \footnotesize
    \label{eqn:surface2}
    S(p)=\left( \dfrac{(p_x+o_u)I_D(p,t_0)}{f_x},\dfrac{(p_y+o_v)I_D(p,t_0)}{f_y},I_D(p,t_0)\right)^{T}
\end{equation}

The {\em principal point} refers to the point on the image plane where the line passing through the center of the camera
lens (optical axis) intersects the image plane. 
The focal lengths in the $x$ and $y$ directions are represented by $f_x$
and $f_y$, respectively.  Each point in the {\em point cloud}
corresponds to a specific position in space and is specified by
its $x$, $y$ and $z$ coordinates. 
To combine multiple such point cloud data structures, we need to transform them to a {\em common coordinate system}. 
This is done using RGB-D Odometry described in Section~\ref{subsubsec:RGBDodo}.

\subsubsection{Point Cloud Fusion}
\label{subsubsec:pcf}
\begin{figure}[!htb]
	\centering
	\subfloat[Point cloud from a single depth image]{
		\includegraphics[width=0.45\columnwidth]{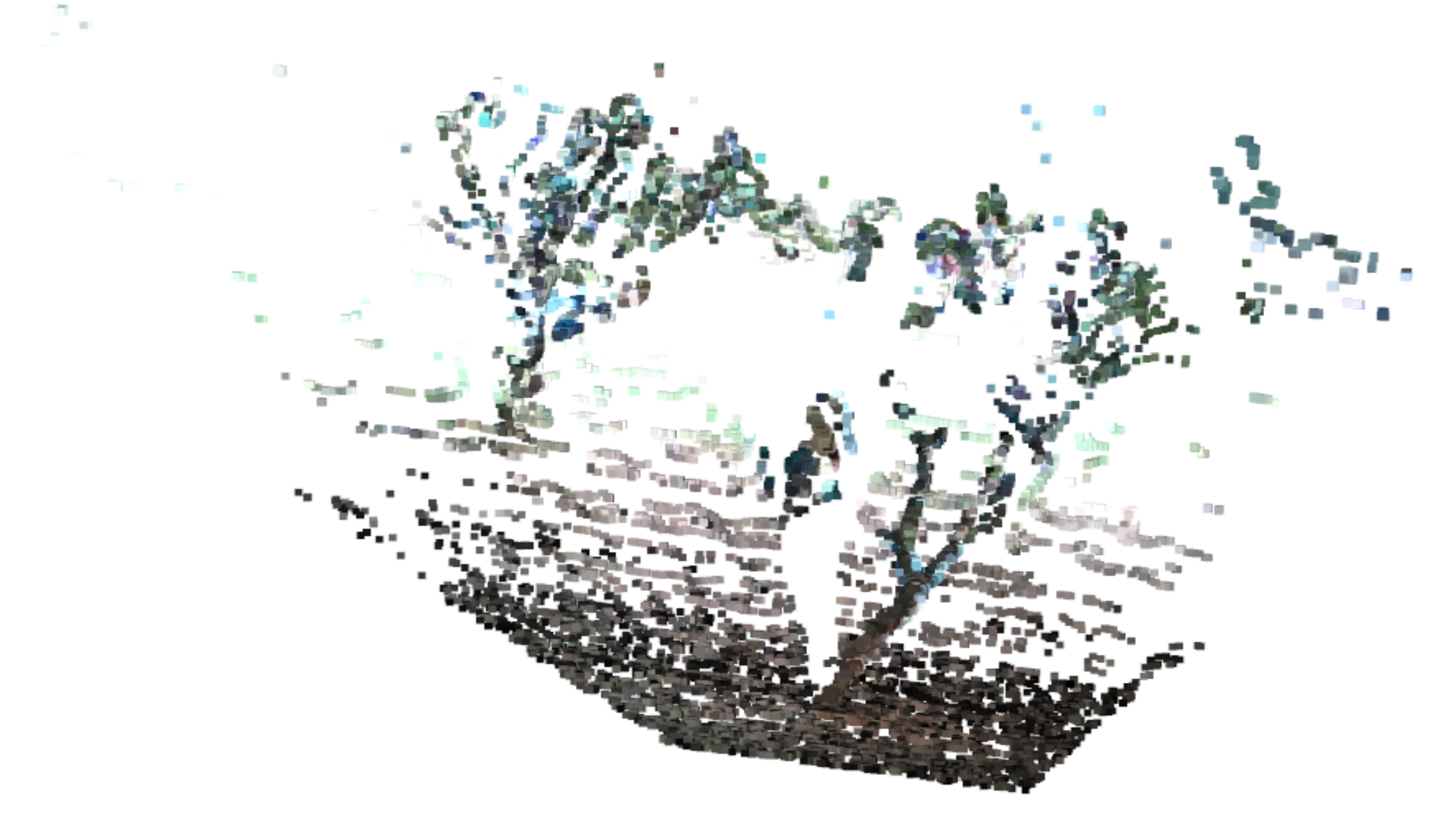}
		\label{fig:pointcloud}
	}
	\subfloat[Fused point cloud]{
		\includegraphics[width=0.45\columnwidth]{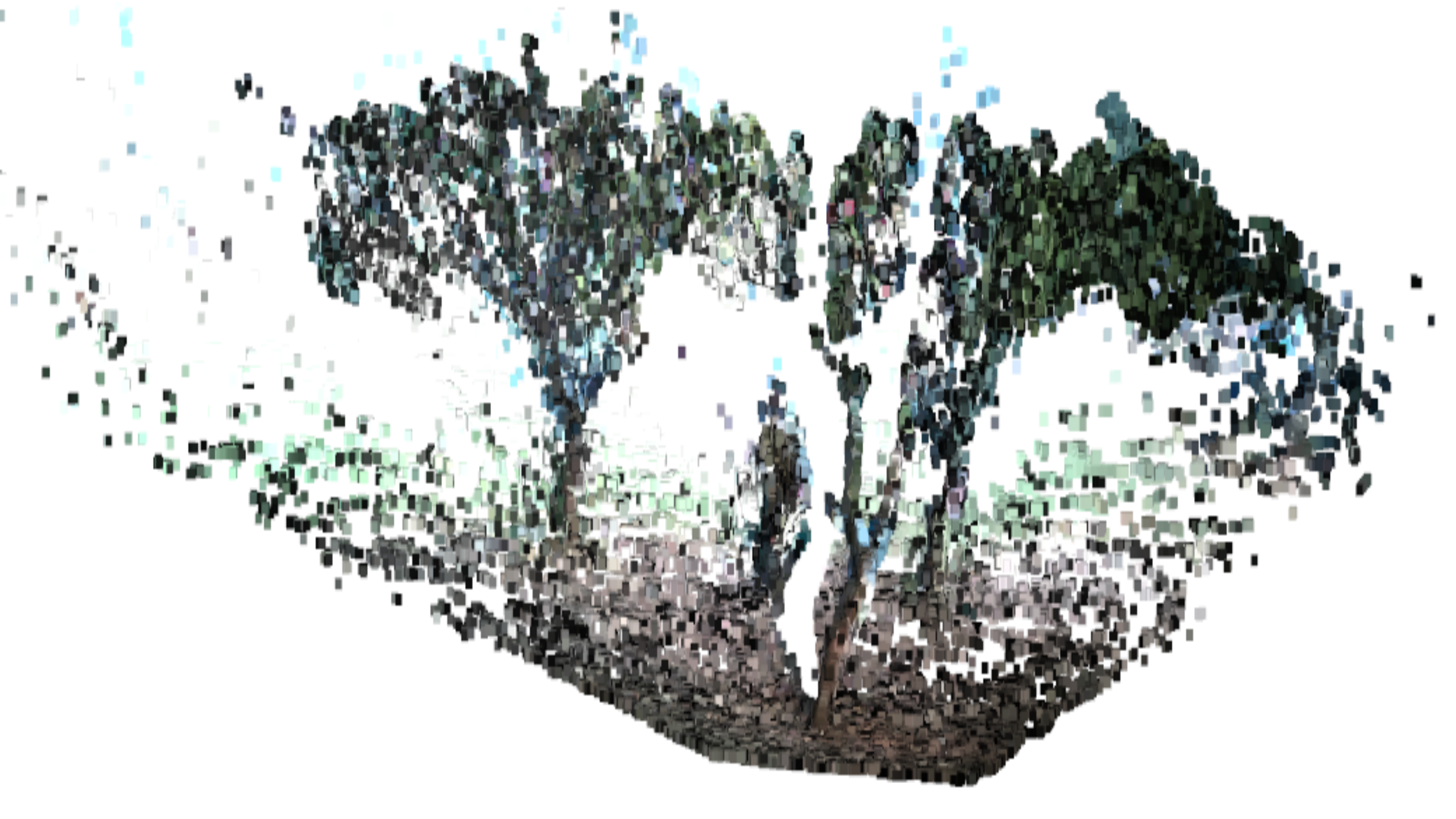}
		\label{fig:voxelfused}
	}
	\caption{Fused point cloud generated after the fusion step. 
    The fusion process fills up holes in the point cloud.}
\end{figure}

Point cloud {\em fusion} is the process
of aligning and merging point clouds from multiple viewpoints (or sensors) to
build a more complete and accurate 3D model of the scene. RGB-D odometry, described in
Section~\ref{subsubsec:RGBDodo}, is used to align multiple point clouds to a common coordinate system. 
Figures \ref{fig:pointcloud} and \ref{fig:voxelfused} show how multiple point clouds are
merged together to form a more complete and less noisy representation of the scene. In our method, we add an
additional step that involves
{\em reprojecting} the merged point cloud back to a 2D image to create
the (\underline{the template}), by following the reverse sequence of
steps (reverse of
Equation~\ref{eqn:surface2}).

%% file: characterization.tex
\section{Characterization}
\label{sec:Characterization}

\subsection{Overview}
\label{subsec:characteroverview}
In this section, we discuss our experimental setup, equipment and datasets used, as well as some
experiments to motivate the problem statement and support the methodology. 
We want to characterize four specific aspects of our setup: 
\fbox{RQ1} The impact of {\em algorithmic noise} on the depth image quality,
\fbox{RQ2} The impact of the {\em frame dimensions} on the computation time and image quality, 
\fbox{RQ3} The performance of different methods to reduce the sparsity by {\em inpainting} depth images, and
\fbox{RQ4} The efficacy of {\em motion compensation} methods as compared to 
a state of the art method namely {\em optical flow}.
For designing our system, these were our most important decision variables.

\subsection{Experimental Setup}
\label{subsec:setup}

\subsubsection{Computational Setup}
We designed our system for the NVIDIA Jetson Nano B01 development Kit, 
a compact and affordable single-board computer created by
NVIDIA. This board is specifically tailored for AI applications,  machine learning, robotics, and
edge computing. 
The technical specifications of our setup are shown in Table~\ref{tab:nanospec}.

\begin{table}[!htb]
  \centering
  \footnotesize
  \caption{Jetson Nano specifications}
  \label{tab:nanospec}
  \begin{tabular}{|l|l|}
  \hline
  \multicolumn{2}{|c|}{\textbf{Processing}} \\ \hline
  GPU & NVIDIA Maxwell architecture\\
  & with 128 NVIDIA CUDA® cores \\ \hline
  CPU & Quad-core ARM Cortex-A57 MPCore processor \\ \hline
  \multicolumn{2}{|c|}{\textbf{Memory}} \\ \hline
  Memory & 4 GB 64-bit LPDDR4, 1600MHz, 25.6 GB/s \\ \hline
  Storage & 16 GB eMMC 5.1 \\ \hline
  \multicolumn{2}{|c|}{\textbf{Camera and Connectivity}} \\ \hline
  Camera & 12 lanes (3x4 or 4x2)\\ 
  &MIPI CSI-2 D-PHY 1.1 (1.5 Gb/s per pair) \\ \hline
  Connectivity & Gigabit Ethernet \\ \hline
  \multicolumn{2}{|c|}{\textbf{Mechanical}} \\ \hline
  Mechanical & 69.6 mm x 45 mm, 260-pin edge connector \\ \hline
\end{tabular}
\end{table}
For comparison we also include some latency related results generated on a workstation computer. The workstation contains an Intel Xeon(R) Gold CPU clocked @ $2.90$GHz with 32 threads. This system is also equiped with $256$ GB of RAM and an Nvidia RTX A4000 GPU.
  
\subsubsection{Real-World Dataset}
We obtain our real-world depth images (RGB-D) using the Intel RealSense D455 depth camera\cite{keselman2017intel}.
To address the shortage of real-world outdoors depth datasets employing stereo cameras, we created our own dataset with manually derived ground truths.
For the technical specifications of the camera refer to Table~\ref{tab:realsense}.

\begin{table}[!htbp]
  \centering
  \footnotesize
  \caption{Intel RealSense camera specifications}
  \label{tab:realsense}
  \begin{tabular}{|l|l|}
  \hline
  \multicolumn{2}{|c|}{\textbf{Operational Specifications}} \\ \hline
  Depth Accuracy & $<$ 2\% at 4m \\ \hline
  Depth Resolution and FPS & 1280x720 up to 90 FPS \\ \hline
  Depth Field of View & $86^{\circ} \times 57^{\circ}$ \\ \hline
  \multicolumn{2}{|c|}{\textbf{Components}} \\ \hline
  RGB Sensor & Yes \\ \hline
  Tracking Module & Yes \\ \hline
  \multicolumn{2}{|c|}{\textbf{Module Specifications}} \\ \hline
  Dimensions & 124mm x 29mm x 26mm \\ \hline
  System Interface Type & USB 3.1 \\ \hline
  \end{tabular}
  \end{table}
 
We created the dataset by shooting at two different locations.
\footnote{The anonymity of the individuals depicted in the dataset photographs, as well as the scenes, has been safeguarded by making
the faces indiscernible.} 
We subsequently used the Realsense SDK to capture the stream of data coming from the camera. This stream 
includes dense $16$-bit integer depth images, a stereo pair of colored images, IMU (Inertial Measurement Unit) and accelerometer data.
The Realsense SDK additionally comes with a built-in hole-filling module~\cite{grunnet2018depth}. 
Finally,
we manually created a set of ground truth images from the raw depth images to evaluate our methods.
 
\subsubsection{Synthetic Dataset}
\begin{figure}[!htb]
  \centering
\subfloat[LN]{
  \includegraphics[width=0.4\columnwidth]{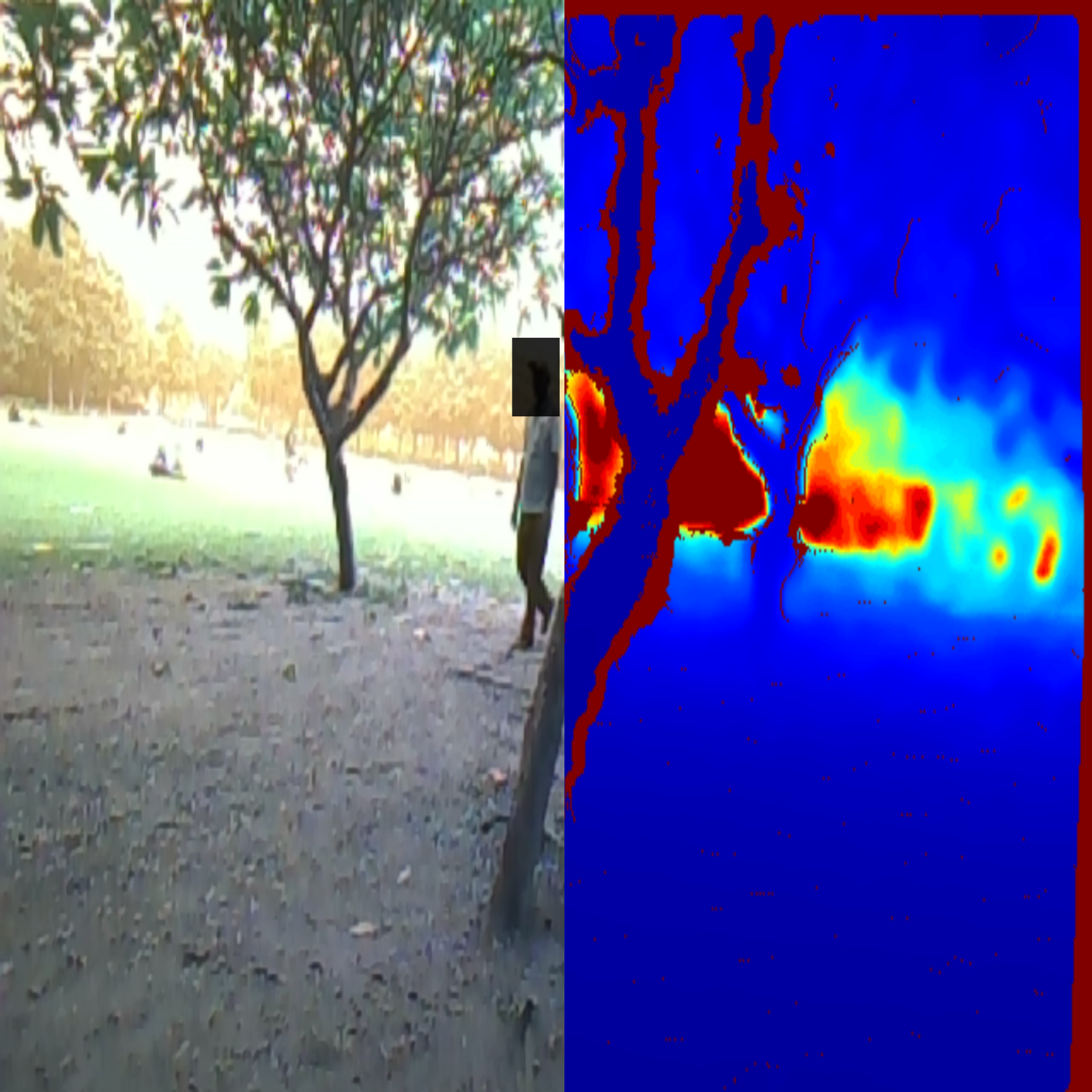}
}
\subfloat[MB]{
  \includegraphics[width=0.4\columnwidth]{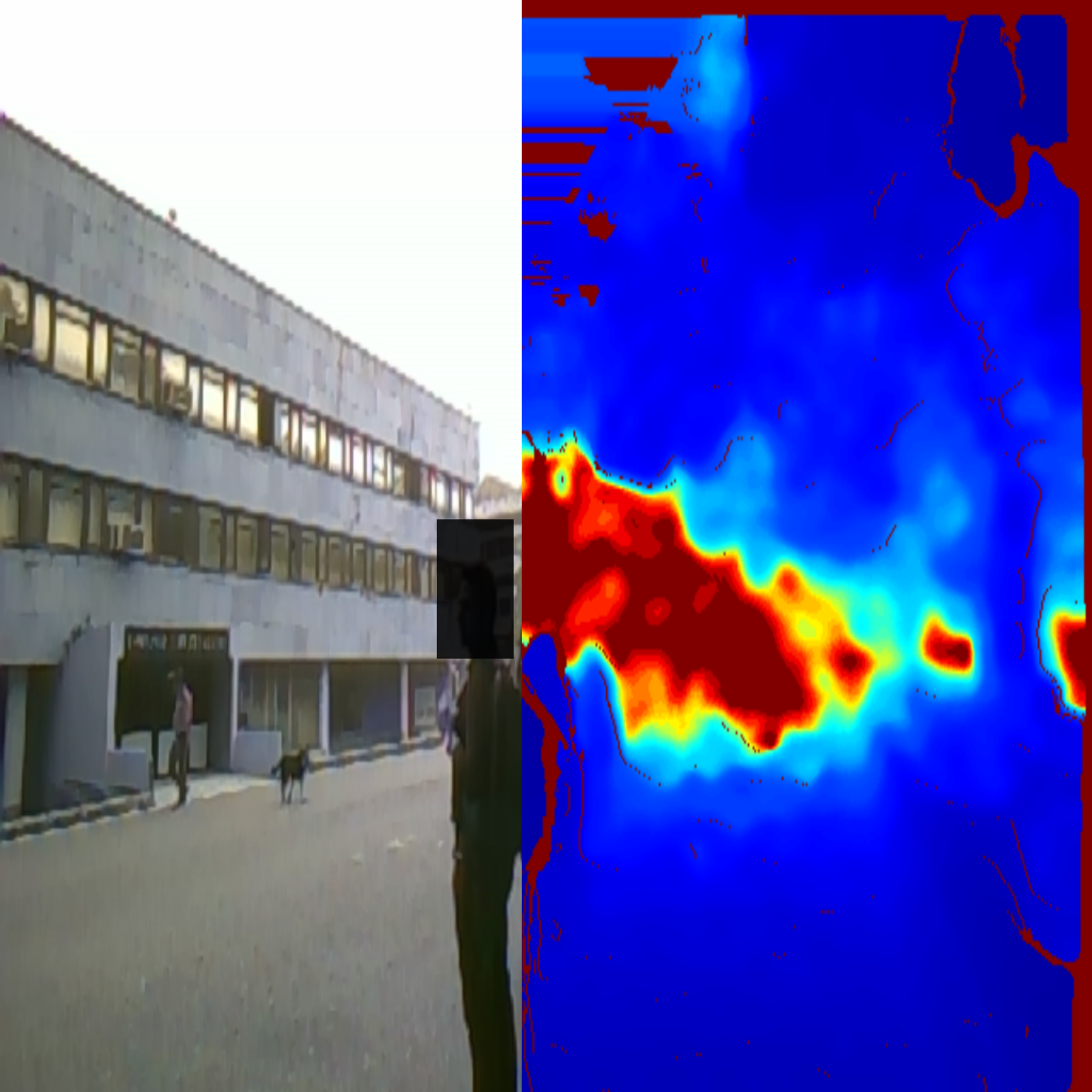}
}
\vspace{-4mm}
\subfloat[KTS]{
  \includegraphics[width=0.4\columnwidth]{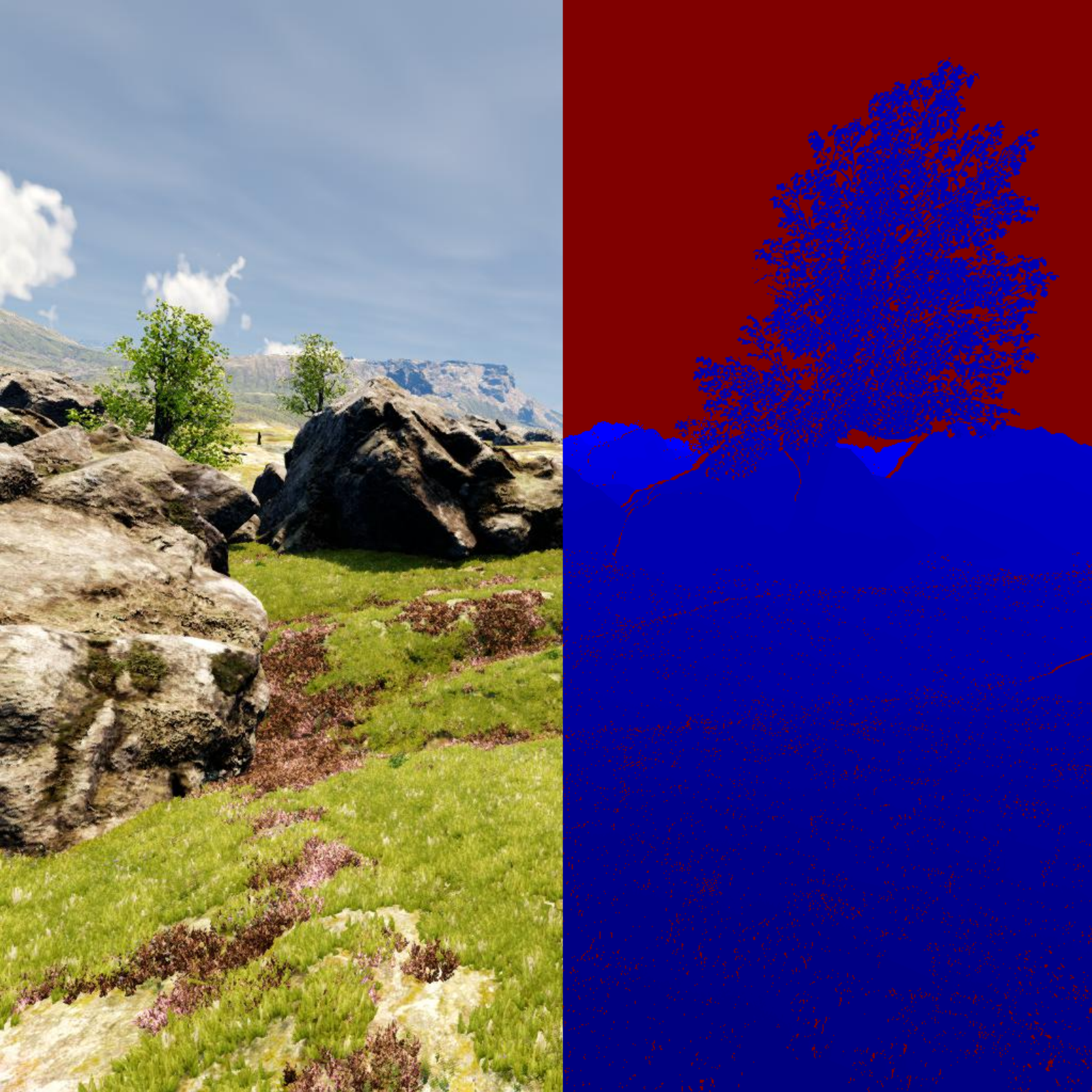}
}
\subfloat[PLF]{
  \includegraphics[width=0.4\columnwidth]{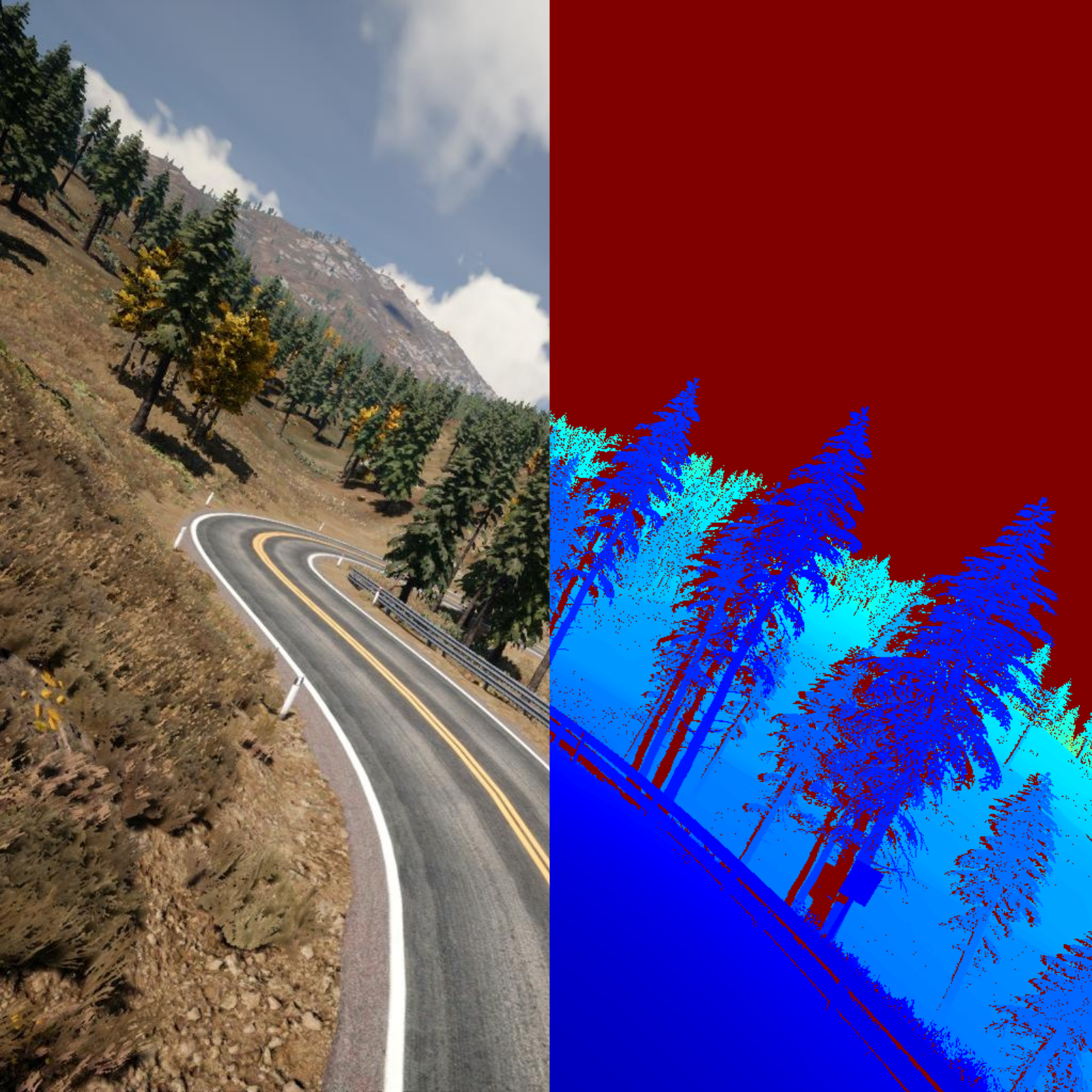}
}
\vspace{-4mm}
\subfloat[KTC]{
  \includegraphics[width=0.4\columnwidth]{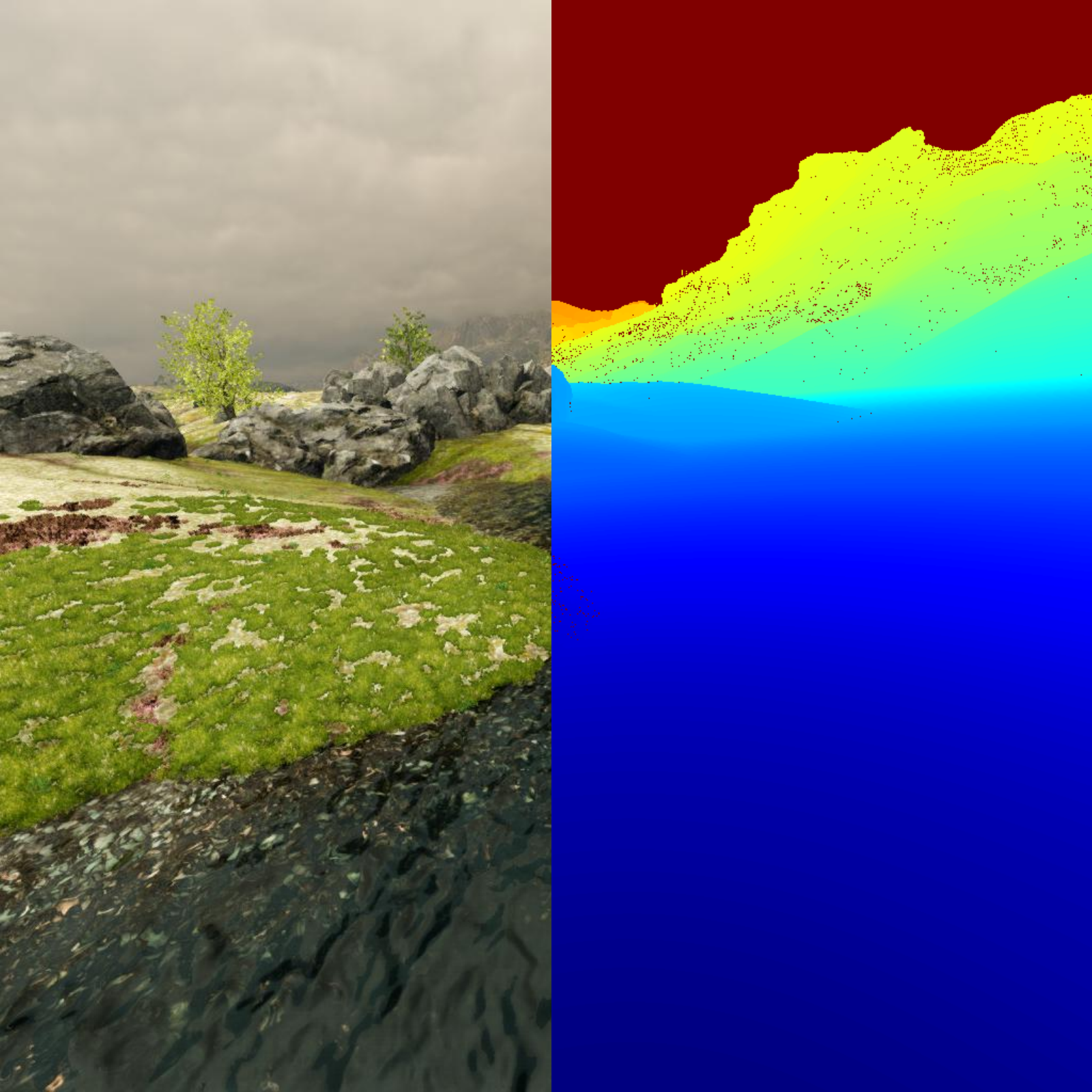}
}
\subfloat[PLW]{
  \includegraphics[width=0.4\columnwidth]{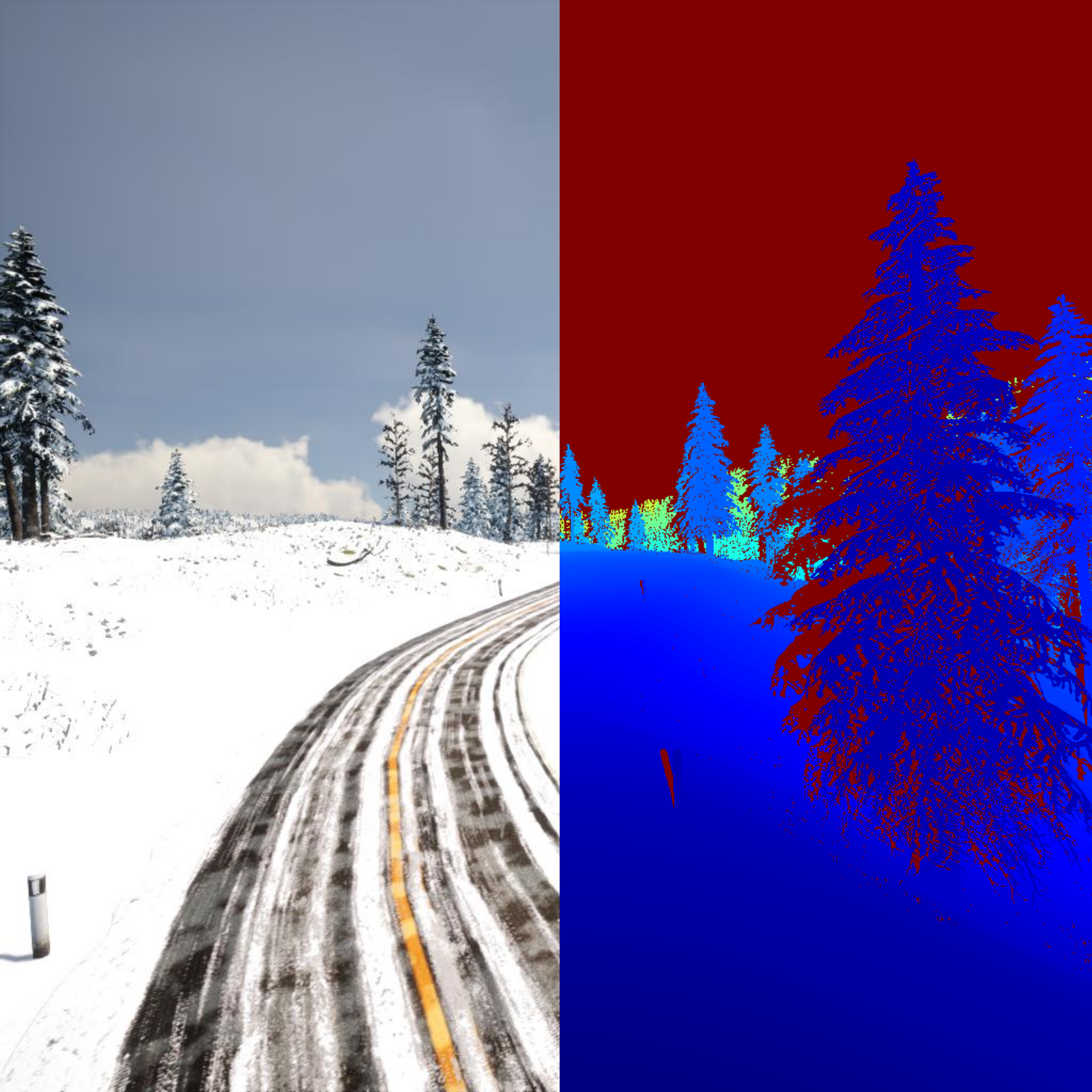}
}
\caption{Depth images from different datasets used in this work}
\label{fig:datasets}
\end{figure}

Moreover, we also used synthetically generated images from 
the Mid-air~\cite{fonder2019mid} dataset, which contains data
from several trajectories of a low-altitude drone 
in an outdoor environment in four different weather conditions generated using the 
Unreal game engine.
The specific datasets used in this work are shown in Table \ref{tab:datasets} 
and example frames are shown in Figure
\ref{fig:datasets}. 
The data includes accurate depth images, stereo pairs of colored images, stereo disparities, stereo
occlusion, surface normals and inertial measurements. 

The reason for using an additional synthetic dataset is as follows.
Developing and testing physical autonomous robots such as drones is costly due to lengthy development cycles and
potential hardware damage during field testing. Prior to real-world experimentation, it is a 
common practice~\cite{} to test the algorithms
using simulators such as AirSim (built on Unreal Engine) ~\cite{shah2018airsim}.
Algorithms are also designed such that they work well for both real-world and synthetic data. 
In line with this philosophy, we also use a hybrid dataset (real-world + synthetic).

\begin{table}[!htb]
  \footnotesize
  \begin{center}
    \begin{tabular}{|l|l|l|l|}
      \hline
     {\textbf{Abbr.}} &{ \textbf{Name}} &  {\textbf{Depth}} & \textbf{Acquisition}    \\
      & &  {\textbf{Dimensions}} & \textbf{Method}    \\
      \hline

    \textit{LN} & Lawns & 360x640 & Realsense   \\
      \hline
    \textit{MB} &  Building & 360x640 & Realsense   \\
      \hline
      
    \textit{KTS} &  Kite Sunny & 1024x1024  & Unreal Engine\\
      \hline
    \textit{KTC} &  Kite Cloudy  & 1024x1024 &Unreal Engine\\
      \hline
    \textit{PLF} &  Procedural Landscape Fall & 1024x1024  & Unreal Engine\\
    \hline
  \textit{PLW} &  Procedural Landscape Winter & 1024x1024 &Unreal Engine\\
    \hline
    \end{tabular}
   \end{center}
   \caption{Datasets used in this work}
    \label{tab:datasets}
  \end{table}

\subsection{RQ1: Impact of Algorithmic Noise}
\label{subsec:impact}

\begin{figure}[!htb]
  \centering
  \subfloat[Average percentage of incorrect pixels in depth images across datasets]
  {\includegraphics[width=0.80\columnwidth]{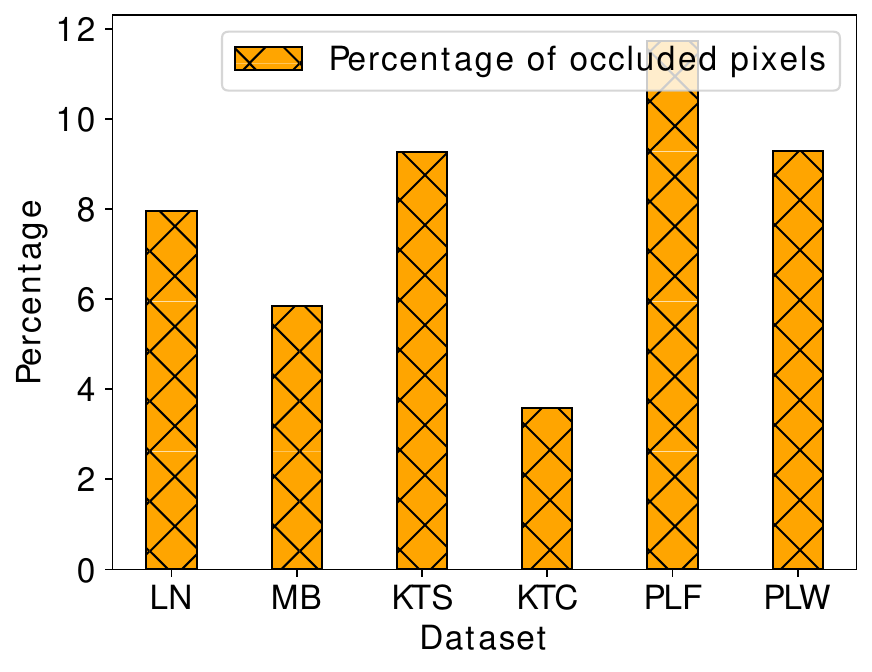}
  \label{subfig:holedist}}
  \\
  \subfloat[Percentage of occluded pixels vs PSNR (dB)]
  {\includegraphics[width=0.80\columnwidth]{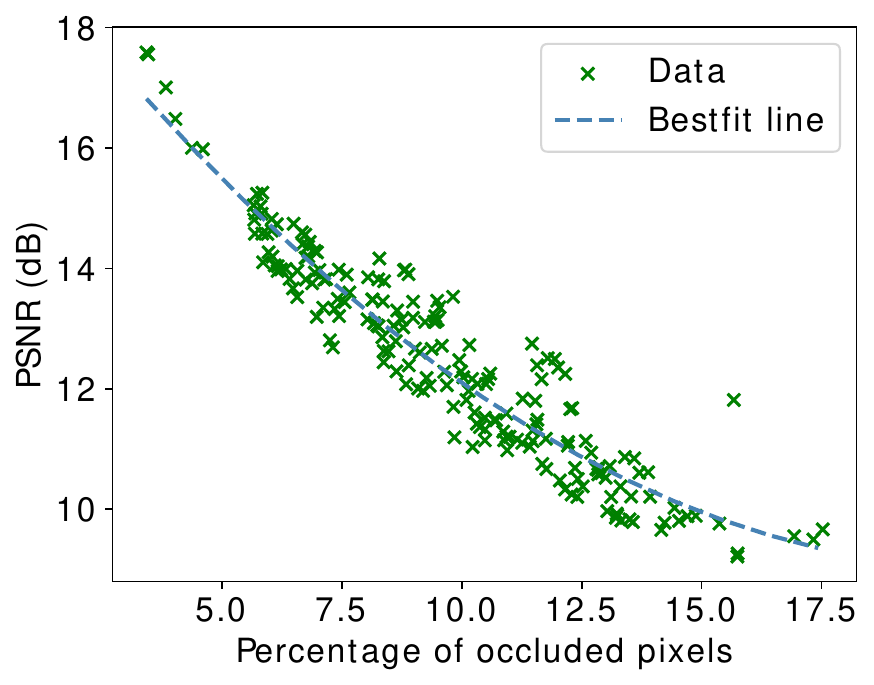}
  \label{subfig:character}}
\caption{Plots depicting (a) the ratio of inaccurate pixels in different datasets and (b) the effect of occlusion holes on the quality of depth images -- PSNR of the raw depth image (vis-a-vis the ground truth)}
\label{fig:occplot}
\end{figure}

The purpose of this subsection is to gain a better understanding of the impact that algorithmic noise has on depth images.  
For the purpose of testing the generalizability of methods in this space, 
we make use of datasets that contain varying degrees of noise as a result of occluded patches
(refer to Figure \ref{subfig:holedist}). Recall that such occlusions lead to algorithmic noise.

The quality of the generated depth images, measured in PSNR (or Peak-Signal to Noise Ratio), decreases significantly as the percentage of noisy pixels in the depth image increases as depicted in Figure \ref{subfig:character}. 
We observe a superlinear 
decrease in the image quality as we increase the percentage of occluded pixels. For a $10\%$ increase in the number of occluded pixels we see a $28.1\%$ reduction in quality in terms of PSNR.

\subsection{RQ2: Impact of the Frame Size on the Registration Latency and Quality}
\label{subsec:resize}

\begin{figure}[!htb]
  \centering
  \subfloat[Quality of registration vs frame latency]
  {\includegraphics[width=1.0\columnwidth]{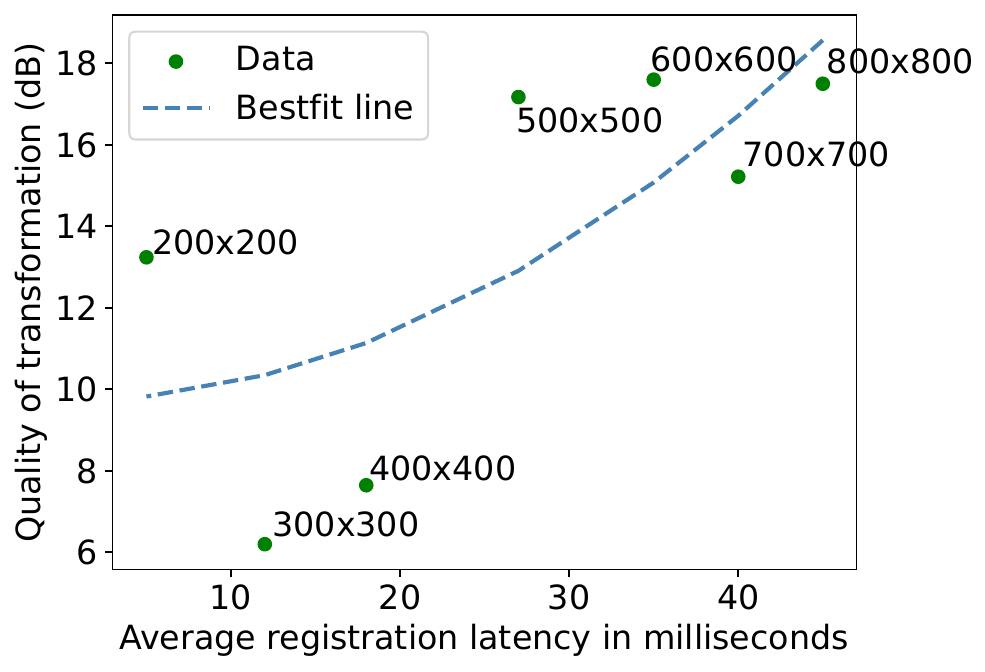}
  }
  \\
  \subfloat[Average quality of the corrected depth images vs frame latency]
  {\includegraphics[width=1.0\columnwidth]{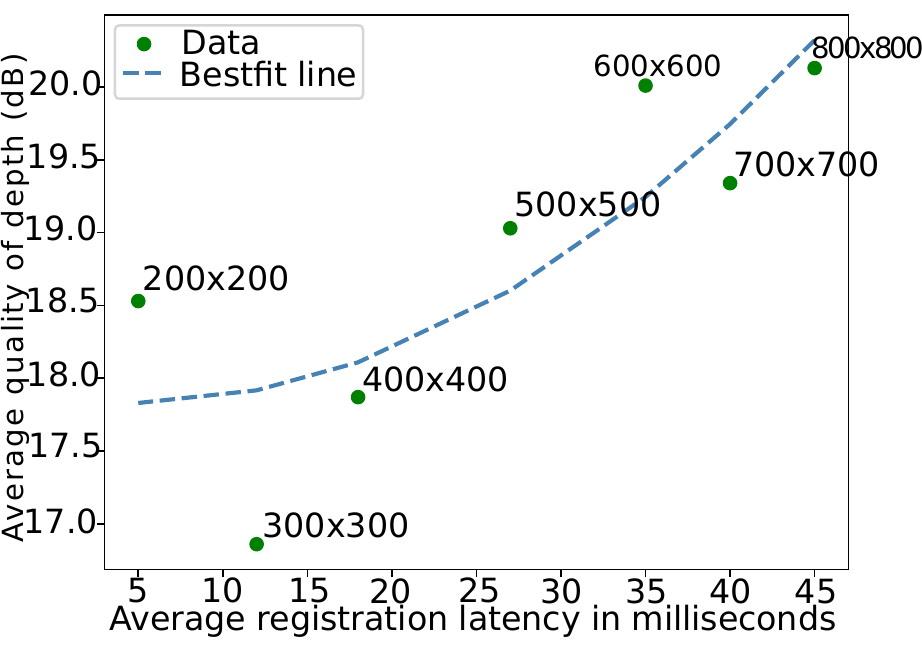}
  }
  \caption{Quality of registration vs the latency. The annotation against each data-point depicts the dimensions of the resized frame.}
  \label{fig:resize}
\end{figure}

To speed up the process of image registration, we can reduce 
the dimensions of the image. This way, the matching
algorithm can find matching pixels in frames faster since the 
search area reduces with a reduction in image dimensions.
This comes at the cost of a lower accuracy in the 
image registration process.
In Figure~\ref{fig:resize}, we show this relationship
graphically. 

We measure the quality of registration
as the PSNR between the original frame 
and the transformed frame. We also plot the average quality
of corrected depth images against the respective latencies associated 
with performing registration on the resized images. We
observe a trade-off between the image quality and the latency that  
broadly shows a 
monotonically increasing trend with the frame size. Occasional
deviation from a monotonically increasing trend can be attributed to 
image-specific variations, noise induced due to the 
data itself and camera motion.  The figure suggests that the best
frame size is $200 \times 200$ (good quality with low latency). This is what
we choose for our experiments. One may always argue that this is an artefact
of our experiment and datasets. We tested with all kinds of images, and the
broad conclusion is that a smaller resolution is good enough from a quality 
perspective given that preservation of low-frequency features tends to affect
image registration the most (also observed in\cite{tzimiropoulos2010robust}). Needless to say, smaller
frame sizes are always desirable from a latency perspective. Hence, even with
other datasets the conclusion is not expected to be very different.

\subsection{RQ3: Inpainting Methods}
\label{subsec:upsamplemethod}
\begin{figure}
    \centering
    \subfloat[Convex upsampling]{
    \includegraphics[width=0.40\columnwidth]{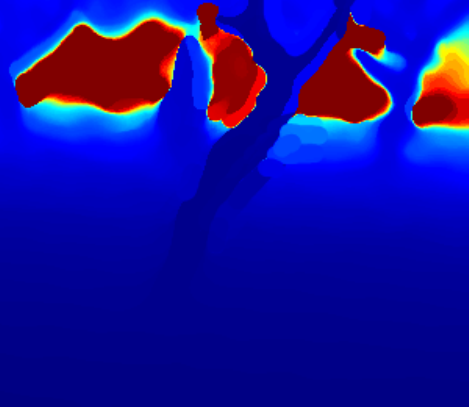}
    \label{subfig:convexupsample}
    }
    \subfloat[Bilninear interpolation]{
    \includegraphics[width=0.4\columnwidth]{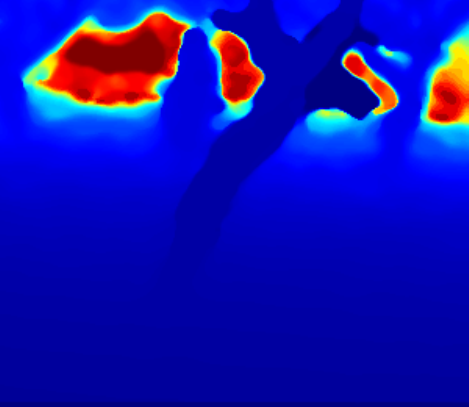}
    \label{subfig:bilinearupsample}
    }
    \\
    \subfloat[Grayscale dilation]{
    \includegraphics[width=0.40\columnwidth]{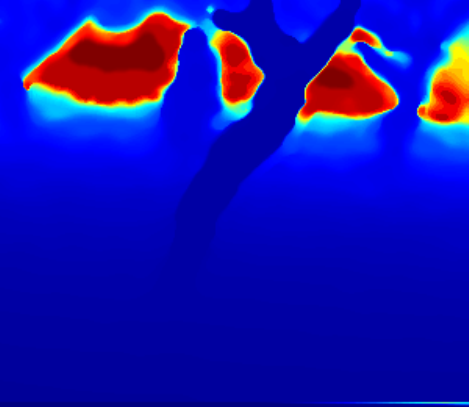}
    \label{subfig:erodeupsample}
    }
    \subfloat[Ground truth]{
    \includegraphics[width=0.40\columnwidth]{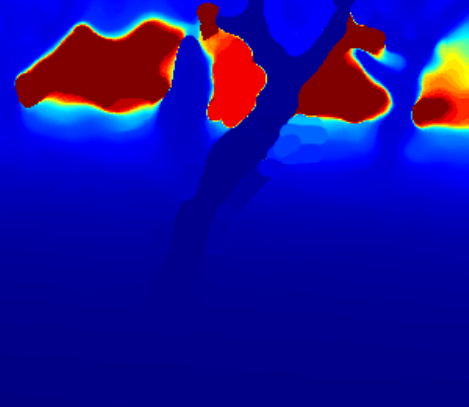}
    \label{subfig:groundupsample}
    }
    \caption{Visual comparison of inpainting methods on depth images.}
    \label{fig:compareupsampling}
\end{figure}

\begin{figure}
    \centering
    \subfloat[Sparse fused depth image]{
    \includegraphics[width=0.40\columnwidth]{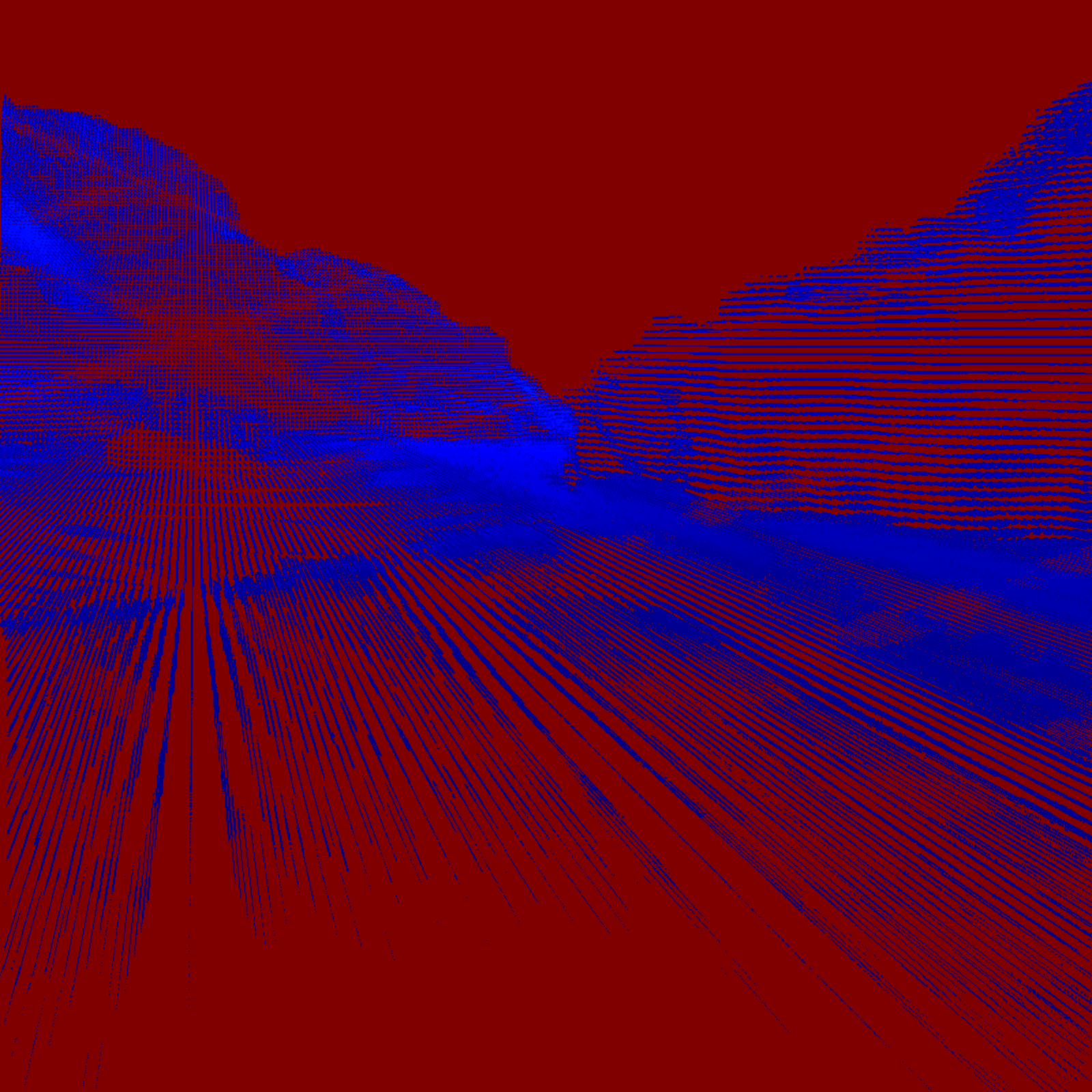}
    \label{fig:voxdownsampled}
    }
    \subfloat[Inpainted depth image]{
    \includegraphics[width=0.40\columnwidth]{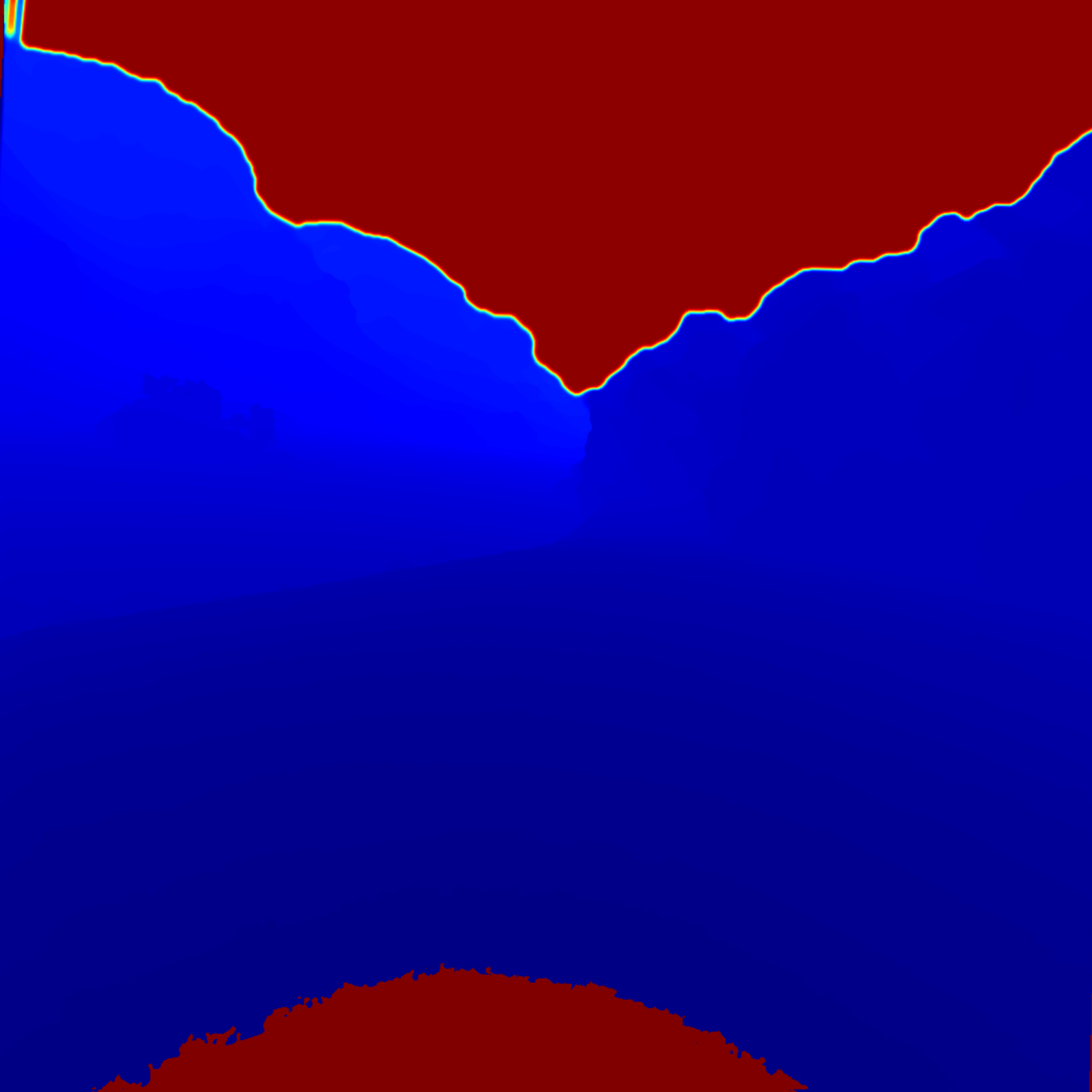}
    \label{fig:voxupsample}
    }
    \caption{Inpainting 2D projections of a fused point cloud to create a dense-scene representation.}
    \label{fig:voxupsample2}
\end{figure}

The two dimensional projection of a point cloud is a sparse depth image as shown in Figure \ref{fig:voxdownsampled}. To create a noise-free 
dense scene representation, the sparse projection would need to be inpainted. 
Conventional image processing methods such as bilinear interpolation
are fast but fail to maintain the sharpness of edges. In certain situations, they may also create unusual visual defects 
as illustrated in Figure~\ref{subfig:bilinearupsample}. Contemporary
inpainting methods that use learning such as convex upsampling~\cite{ferstl2013image}, yield nearly flawless outcomes (see Figure~\ref{subfig:convexupsample}).
However, they are prohibitively slow. 

We implemented three methods for the purpose of comparison: 
bilinear interpolation\cite{sa2014improved}, convex upsampling method proposed by
Teed et al.~\cite{teed2020raft}, and grayscale dilation\cite{vincent1993morphological}.
Grayscale dilation is a local image filter 
that is used for noise reduction in image processing tasks where preserving structural 
information is important.
The quality of generated images and associated 
latencies on the Jetson Nano board are shown in Table~\ref{tab:upsamplingmethods}.
We find that the latency associated with 
grayscale dilation falls within the acceptable threshold for producing
corrected frames at $20$ FPS; it performs better than bilinear interpolation.

\begin{table}[!htb]
    \centering
    \begin{tabular}{|l|l|l|}
         \hline
         \textbf{Method} & \textbf{PSNR} (dB) & \textbf{Latency} (ms) \\
         \hline
          Bilinear interpolation & 16.35 & 8\\
          \hline
          Convex upsampling & 34.39 & 237\\
          \hline
          Grayscale dilation & 17.46 & 94 \\
          \hline
    \end{tabular}
    \caption{Comparison of the image quality and latencies of different inpainting methods.} 
    \label{tab:upsamplingmethods}
\end{table}

\subsection{RQ4: Relevance of the State-of-the-art: Optical Flow based Techniques}
\label{subsec:opticalflow}

An extremely popular method of depth correction is motion compensation using optical flows
~\cite{avetisyan2016temporal,matyunin2011temporal}. An {\em optical flow}
enables the system to predict the movement of
specific objects in the scene at a pixel level, 
making it more amenable to motion correction as compared to image
registration. Here, we outline our arguments \underline{opposing}
the utilization of optical flows. 
Although several contemporary
embedded systems such as the Jetson AGX Orin~\cite{jetsonagxorin} possess a built-in optical flow accelerator, but
they are limited by other shortcomings such as high power consumption, weight and cost. The 
Orin has a price tag of
approximately $\$2,000$ USD and a weight of approximately $1.4$ kg. This exceeds the constraints that we 
set in Section~\ref{sec:Introduction}.

We tested the RAFT optical flow network~\cite{teed2020raft}
with the necessary optimizations and a CUDA implementation
of the Horn-Schunck optical flow technique~\cite{mizukami2007optical} on our Jetson Nano board. 
We observed that 
RAFT network requires $15.5$ seconds to perform one inference, while the Horn-Schunck approach takes $2.45$ seconds to
compute the optical flow for a single frame. Both of these latencies significantly exceed the acceptable threshold for a
real-time system, where each frame needs to be computed/processed within
$50$ milliseconds (20 FPS). This proves that optical flow based
systems are not a suitable choice for real-time applications of this nature.

\begin{tcolorbox}[colback=blue!10!white,colframe=blue!60!black,title=Characterization Insights]
  \fbox{RQ1:} There is a super-linear decrease in the depth image
   quality with an increase in the percentage of occluded pixels.\\
   \fbox{RQ2:} The quality of image registration goes down when frame dimensions are lowered, but latency goes down as well. The ideal resolution (in our experiments)
    was found to be $200 \times 200$ because it had $73 \%$ better registration quality than higher resolutions while also being $3 \times$ faster.\\
  \fbox{RQ3:} Grayscale dilation as a depth inpainting technique preserves edges and is faster than learned methods.\\
  \fbox{RQ4:} Optical flow based motion compensation techniques are too slow to be implemented on real time embedded system applications.\\
  \end{tcolorbox}

%% file: implementation.tex
\section{Implementation}
\label{sec:Implementation}

\subsection{Overview}
\label{sec:overview}

\begin{figure*}[!htb]
        \centering
	\subfloat[]
        {
            \includegraphics[width=2.0\columnwidth]{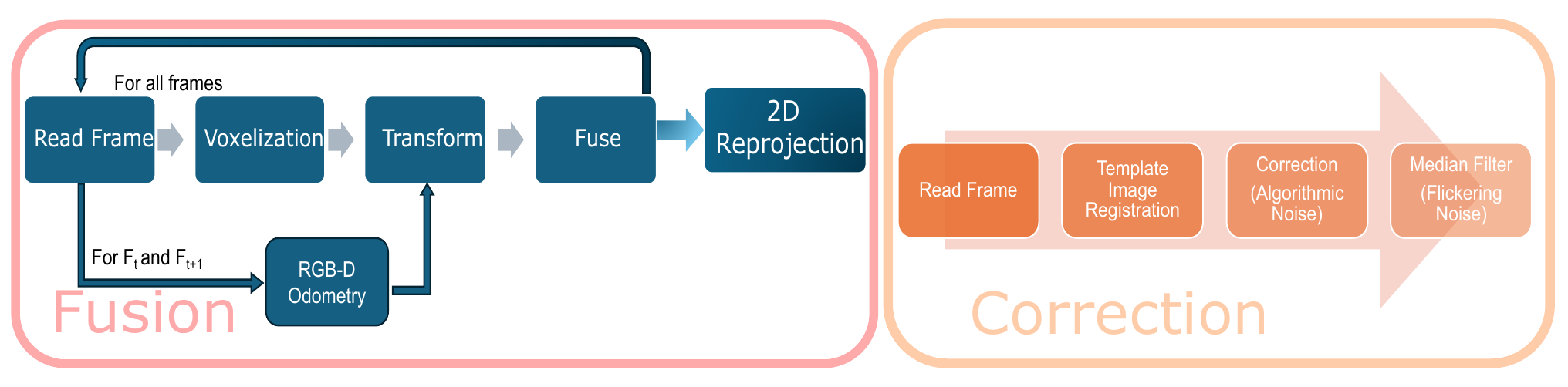}
            \label{subfig:dividedflow}
        }
        \\
        \subfloat[]{
		\includegraphics[width=0.6\columnwidth,height=0.35\columnwidth]{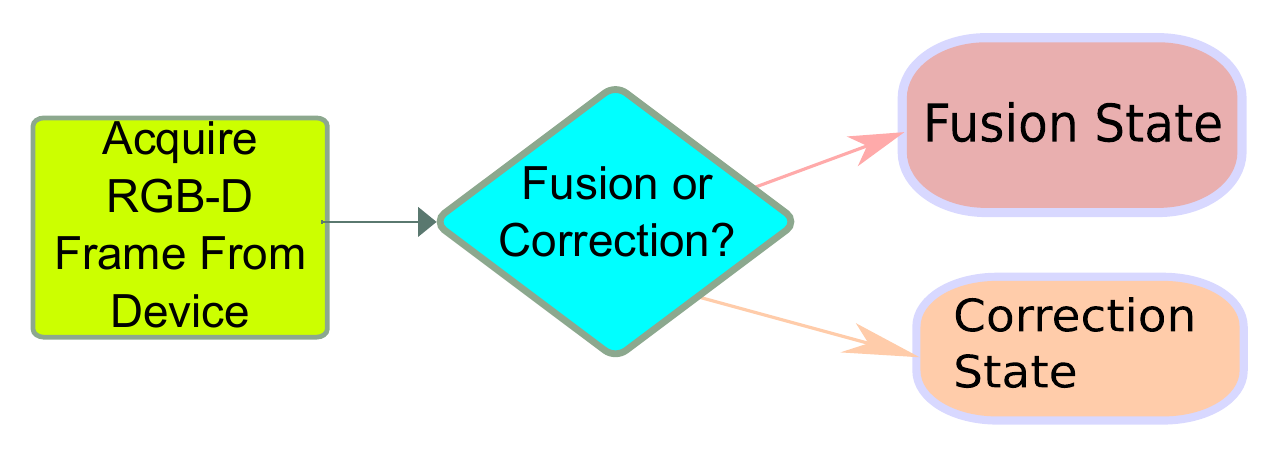}
		\label{subfig:broadflow}
	}
    	\subfloat[]{
		\includegraphics[width=1.4\columnwidth,height=0.45\columnwidth]{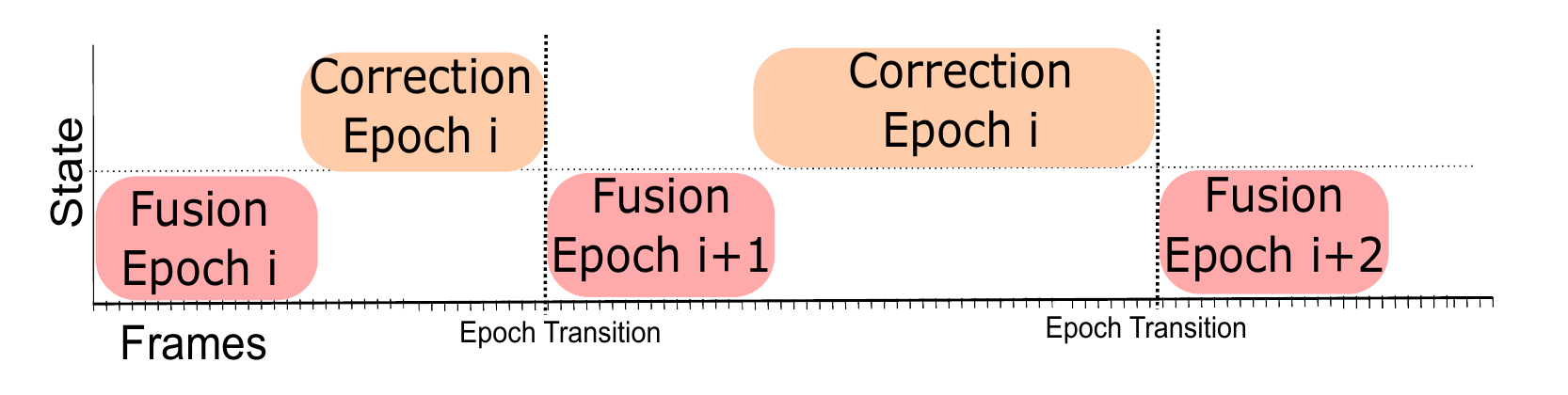}
		\label{subfig:statediag}
	}

	\caption{(a)Flow diagram of the proposed method. The two states of the system have been encased in labeled boxes. The left box, the {\em fusion} state, is described in Section \ref{subsec:scene} and is responsible for creating an accurate scene representation in the form of a template image. The right box, named the {\em correction} state, is described in Section \ref{subsec:depthcorr}, which uses the template image generated in the last state to fix the inaccuracies in the raw depth images.(b) The conditional switching between the states is explained in Section \ref{subsec:adaptive}.(c) A time diagram showing how the epochs proceed. } \label{fig:flow}
\end{figure*}

Given that we have already seen that
occluded regions serve as a significant determinant 
of the depth image quality (in Section~\ref{subsec:impact}), 
we propose an {\em epoch}-based depth correction method to correct these inaccuracies.
An {\em epoch} consists of two states: {\em fusion} and {\em correction}.
In Figure~\ref{subfig:dividedflow}, we depict these two states in the two labelled boxes. 
The left box, representing the fusion state, contains RGB-D 
odometry, transformation and fusion modules. They are  
described in detail in Section~\ref{subsec:scene}. The right box represents the correction state 
which contains the image registration and final correction modules. These have been explained in Section~\ref{subsec:depthcorr}. Finally, we need to automatically 
switch between these two states. This is handled by the {\em epoch} transition module represented in Figure~\ref{subfig:statediag}. This module measures the 
accuracy of image registration to decide the point of state transition. We describe the epoch transition module in Section~\ref{subsec:adaptive}.

Our objective is to rectify the erroneous patches 
generated in the raw depth images and eliminate sensor noise,
while also achieving the target frame rate of 20+ FPS.
The goal of the system is to
exploit spatio-temporal similarities in a scene in three 
dimensions to fill in the gaps in the collected depth images. 
To achieve a high framerate, the system needs to run
high-latency components as infrequently as possible while also 
being adaptive and responsive.

\subsection{Point Cloud Fusion}
\label{subsec:scene}
The first action is to use the first $n$ frames ($F_1 \ldots F_n$) to create a fused
point cloud that represents an accurate 3D representation of the scene.
Figure~\ref{fig:fusiondiagram} shows the process by which a fused point cloud 
is created using the first $n$ frames at the beginning of an epoch.

Our method needs to ensures two properties: \\
\circled{1} It must not exceed the limited $4$GB  memory available on the embedded device. \\
\circled{2} It needs to run in parallel and use all the concurrent resources of the in-built GPU. 

\subsubsection{Voxelization}
\label{subsubsec:Voxelization}

To fuse incoming depth frames, we first transform each incoming color-depth 
image pair (RGB-D image) into a point cloud using Equation~\ref{eqn:surface2}.
Before we can design a fast strategy for fusing point clouds, we need to manage 
the application's memory footprint.
Storing all the points in a fused point clouds can lead to high latencies and inefficient use of memory.
Also, the distance between successive points in the point cloud
is variable -- this makes it hard to 
process it.
We thus store the point cloud using
a discrete three-dimensional grid data structure called a {\em voxel grid}, 
where the constituent points (voxels) represent an equal-sized cubic volume. They are 
uniformly spaced. This conversion process is known as {\em voxelization}. 
Note that in a voxel grid, each
point is associated with a Boolean value; if it is 1, then it means that
the corresponding point 
exists in the point cloud, and vice versa. A voxel grid is thus a sparse 3D matrix.

Since each pixel in the depth image can be processed independently, 
we implement a CUDA~\cite{luebke2008cuda} based parallel algorithm as described in Algorithm \ref{algo:voxel} to implement voxelization.

\subsubsection{RGB-D Odometry}
\label{subsubsec:transform}
We define the interval between the processing of Frames $F_t$ and $F_{t+n-1}$
as the {\em fusion window} (refer to Figure~\ref{fig:fusiondiagram}).
The length of the fusion window 
is typically 10 frames (studied in Section~\ref{subsubsec:winsize}), i.e., $n=10$.
We utilize the visual odometry technique, described in Section~\ref{subsubsec:RGBDodo},
as outlined by Steinbr{\"u}cker 
et al. ~\cite{steinbrucker2011real} to approximate the camera motion for the
frames within the window.
This camera motion estimation is performed only once at the beginning of an {\em epoch},
between frames $F_t$ and $F_{t+1}$. We refer to the difference of these two
frames as the {\em transform}.
In order to minimize the overall latency, we assume that the velocity does not change significantly
for the duration of the fusion window. In other words, the difference between
consecutive frames is approximated to be equal to the {\em transform}. 

\subsubsection{Fusion}
\label{subsubsec:fuse}
\begin{figure}[!htb]
	\centering
  		\subfloat[]{
            \includegraphics[width=1.0\columnwidth]{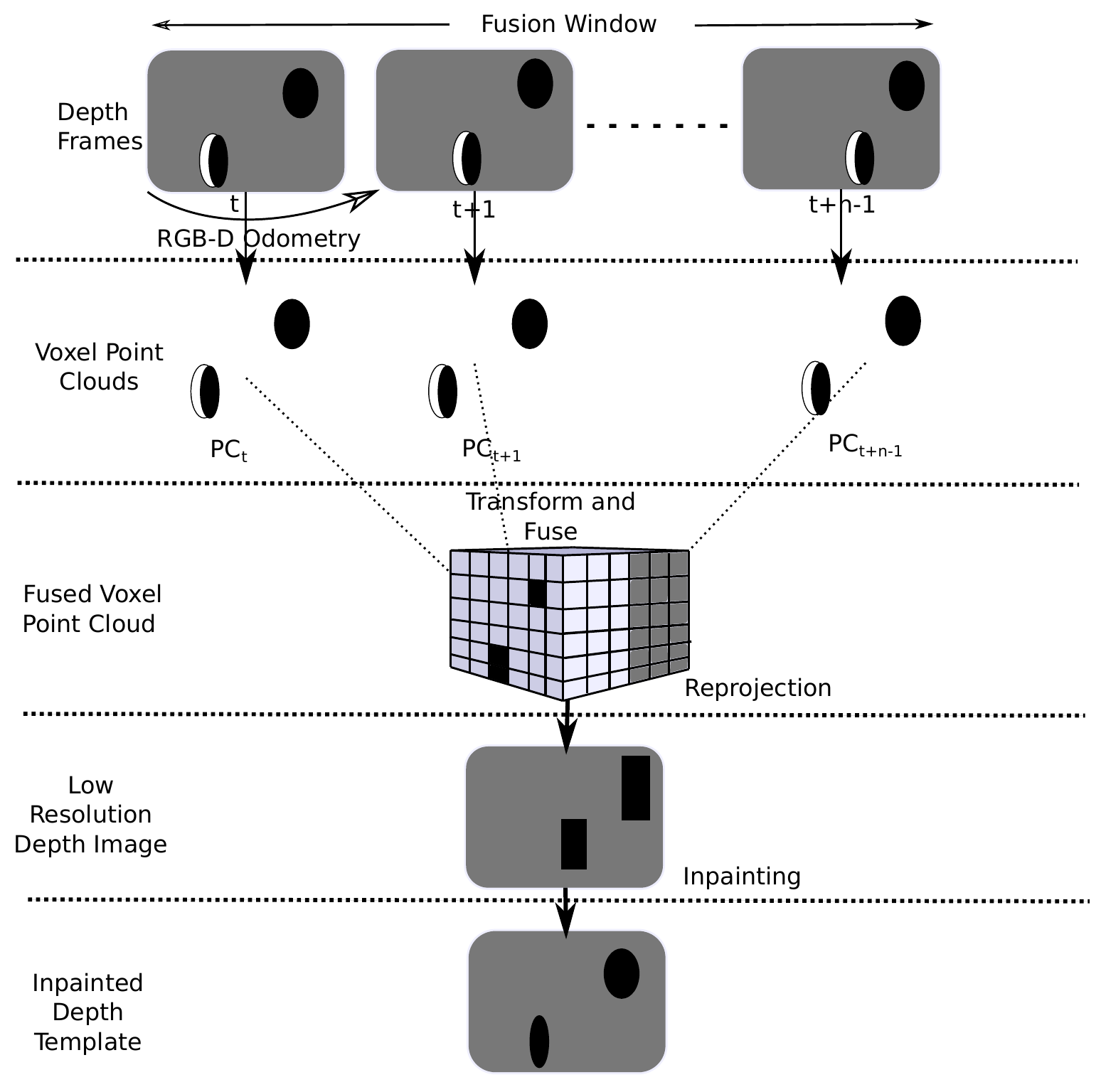}
            \label{fig:fusiondiagram}
            }
            \\
            \subfloat[]{
  		\includegraphics[width=1.0\columnwidth]{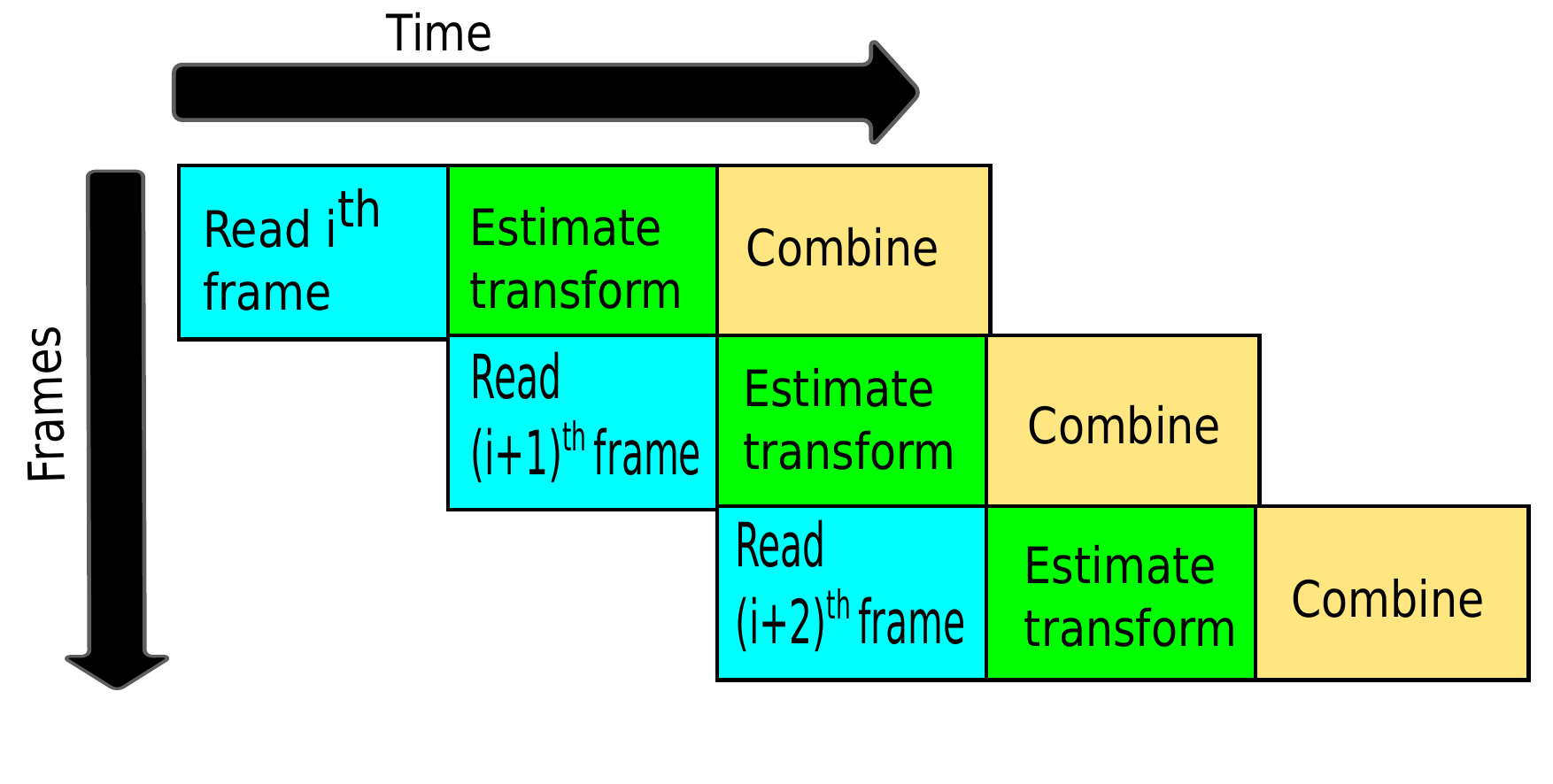}
		\label{subfig:pipeline}
	   }
    \caption{(a) Visual representation of the fusion process~\ref{subsec:scene}. (b) Software pipeline used to ensure the efficiency of depth correction~\ref{subsec:depthcorr}.}
\end{figure}

The final step comprises taking the voxel grid from the previous step and 
integrating it into a common voxelized point cloud. 
The entire process is described in Figure~\ref{fig:fusiondiagram}. 
The integration process uses a Boolean OR operation to determine if a voxel is occupied. A Boolean OR between two voxel grids is defined as 
a voxel wise OR operation. If $V_{(i,j,k)}$ represents a voxel at a discrete location $(i,j,k)$, then the Boolean OR ($\vee $) between two voxel grids
$V_1$ and $V_2$ with dimensions $N \times N \times N$ can 
be defined as:

\begin{equation}
    \label{eqn:voxelor}
    V_1 \vee V_2 = \{ V_{1(i,j,k)}\ OR\    V_{2(i,j,k)} ; 0 \leq i,j,k < N,\forall i,j,k \in \mathbb{Z} \}
\end{equation}

\begin{equation}
    \label{eqn:fuse}
    PC_{final} = f^T(...f^T(f^T(PC_t)\vee PC_{t+1})\vee....) \vee PC_{t+n-1}
\end{equation}

In Equation~\ref{eqn:fuse}, 
the final fused voxel point cloud $PC_{final}$ is generated by successively transforming 
and integrating new point clouds. Here $f^T$ is the transform function.
(estimated in Section~\ref{subsubsec:transform}).
Note that {\em applying the 
transform} means that we
multiply each point cloud pixel with the corresponding one in the transformation matrix.
$\vee$ is the Boolean OR operation between voxel grids.
Note that during the fusion process the system uses the template generated during 
the previous epoch. This leads to a degradation in quality but is still better 
than using raw depth frames.

\subsubsection{2D Projection}
\label{subsubsec:graymorph}
The fusion process results in a {\em fused} 3D voxelized point cloud.
Whereas, the raw depth images are in 2D. 
To bridge the gap, the voxelized point cloud in 
Figure~\ref{fig:voxelfused} is projected back to a 2D $16$-bit integer depth image 
as shown in Figure~\ref{fig:voxdownsampled}. 
This 2D {\em template depth image} is used to correct the raw
2D depth frames.
The 2D depth image generated after fusion is
\underline{too sparse} for it to be used in our correction step.
A sparse template image would lead to misidentification of a depth value in 
the raw frame.
There are local~\cite{kopf2007joint} 
and non-local~\cite{buades2005non} filter-based
methods for inpainting 2D depth images. We chose a class of local techniques called {\em morphological transformations}\cite{vincent1993morphological} to
perform the inpainting due to their simplicity and parallelizability. 
Specifically,
{\em dilation} (Equation~\ref{eqn:dilate}) is a morphological operation where each pixel in a grayscale image is
replaced by the maximum in its neighborhood. 
Similarly, {\em erosion}(Equation~\ref{eqn:erosion}) is a morphological operation where each pixel in a
grayscale image is replaced by the minimum in its neighborhood.

\begin{equation}
    \label{eqn:dilate}
    \begin{split}
        I_{transform}(x,y) & = \text{max}\left(I\left(i,j\right)\right) \text{where} \\ 
        & i \in \left[x-\lfloor N/2 \rfloor,x+\lfloor N/2 \rfloor\right]; \\
        & j \in \left[y-\lfloor N/2 \rfloor,y+\lfloor N/2 \rfloor\right]
    \end{split}
\end{equation}

\begin{equation}
    \label{eqn:erosion}
    \begin{split}
        I_{transform}(x,y) & = \text{min}\left(I\left(i,j\right)\right) \text{where} \\ 
        & i \in \left[x-\lfloor N/2 \rfloor,x+\lfloor N/2 \rfloor\right]; \\
        & j \in \left[y-\lfloor N/2 \rfloor,y+\lfloor N/2 \rfloor\right]
    \end{split}
\end{equation}

This template depth image is used in the depth correction step to fill in the incorrect patches 
in the incoming raw depth images from the stereo camera. 
Once we have our {\em template}, the first part of the {\em epoch} is complete, and we move on to 
the {\em correction} state.

\begin{algorithm}[!htb]
	\footnotesize
    \caption{Voxelization algorithm}
    \label{algo:voxel}
		\hspace*{\algorithmicindent} \textbf{Input:} \\
		\hspace*{\algorithmicindent}$\texttt{CX,CY,FX,FY}$  \Comment{Camera intrinsic params}\\
		\hspace*{\algorithmicindent}$\texttt{GridSize,XVoxSize,YVoxSize,YVoxSize}$ \Comment{Voxel params}\\
		\hspace*{\algorithmicindent}$\texttt{imgW,imgH}$ \Comment{Image dimensions}
	\begin{algorithmic}[1]
	\Function{\textsc{toSurfaceCaller}}{} \Comment{CUDA kernel function}
		\State{$\texttt{xIndex} \gets \textsc{ThreadIdx.x}$} \Comment{One thread per input pixel}
		\State{$\texttt{yIndex} \gets \textsc{ThreadIdx.y}$}
		\If{$(\texttt{imgH,imgW}) \textsc{ within image limits}$}
			\State{$\texttt{z} \gets \texttt{flImgD}[\texttt{xIndex}][\texttt{yIndex}]$}
			\If{$\texttt{z} \neq 0.0$} \Comment{Calculate voxel coordinate for pixel}
				\State{$xVox \gets \textsc{floor}\left(\frac{{\left(\left(\texttt{xIndex} - \texttt{CX}\right) \times \texttt{z}\right)}}{{\texttt{FX} \times \texttt{XVoxSize}}}\right) + \frac{\texttt{GridSize}}{2}$}
				\State{$yVox \gets \textsc{floor}\left(\frac{{\left(\left(\texttt{yIndex} - \texttt{CY}\right) \times \texttt{z}\right)}}{{\texttt{FY} \times \texttt{YVoxSize}}}\right) + \frac{\texttt{GridSize}}{2}$}
				\State{$zVox \gets \textsc{floor}\left(\frac{\texttt{z}}{\texttt{ZVoxSize}}\right)$}
				\If{$(\texttt{xVox,yVox,zVox}) \text{ inside voxel}$}
					\State{$\texttt{voxelGrid}[\texttt{xVox}][\texttt{yVox}][\texttt{zVox}] \gets 1$}
				\EndIf
			\EndIf
		\EndIf
	\EndFunction
	\end{algorithmic}
	\end{algorithm}

\subsection{Depth Correction}
\label{subsec:depthcorr}
The final stage of the process, depth correction, is detailed in this section. 
The following are a few of the challenges we encountered while developing a 
fast depth correction step:  \\
\circled{1} The correction must take into account the robot's 
motion in order to correctly rectify the values of the inaccurate pixels. \\
\circled{2} The latency may increase significantly due a bottleneck caused by reading 
frames from the device (I/O operations). \\
Considering these factors, we created a motion-aware correction module
that is pipelined and employs all of the system's concurrent hardware.

\subsubsection{Template Image Registration}
\label{subsubsec:template Registration}
Let the {\em template} be $I_{T}$. 
When we read a new frame $I_{F}$, we combine it with the template
to create the corrected frame. 
Before we combine the new frame and the template, 
we must first transform $I_{T}$ to the coordinate plane of the current frame. 

This is achieved by estimating the affine transform
described in Section~\ref{subsubsec:imgreg}. We use the 
{\em ORB} (Oriented FAST and Rotated BRIEF) feature matching algorithm
introduced by Rublee et al.~\cite{rublee2011orb}
to match features across a pair of frames and estimate the transform. Since estimating the
transform is a time-consuming task, as we saw in Section \ref{subsec:resize}, 
we downscale the RGB frames to $200 \times 200$ pixel frames,
and then estimate the transform. 

\subsubsection{Algorithmic Noise Correction}
\label{subsubsec:correction}
Finally, the transformed template is combined with 
the foreground depth by replacing the invalid pixels in the 
new image with the corresponding pixels in the template.

\begin{equation}
    \label{eqn:depthcor}
        I_{C} = \{ I_{F}[i,j] \mid \text{if } I_{F}[i,j] \text{ is valid;}  \text{ else } M(I_{T}[i,j]) \}
\end{equation}

Here, $I_{C}$ is the combined final image and $M(I_T)$ is the transformed template image. 
A pixel is considered invalid if either 
\circled{1} its value is less than half of the corresponding pixel's value in the template ($M(I_T)$)  or 
\circled{2} the pixel's value is greater than 
the corresponding pixel's value in the template $M(I_T)$.

\subsubsection{Flickering Noise Filter}
\label{subsubsec:imgfilter}
We apply a median filter across the combined depth image ($I_C$) 
to filter out flickering noise by exploiting the spatial similarities in a depth image.
A median filter is a spatial filter that replaces a pixel value by the median of its neighborhood. 
It is effective in removing \lq salt and pepper\rq type of noise from images while still preserving edges. 
It is most often implemented by moving a square window $W$ centered at a pixel $(x,y)$ across the image. 
The median of all the values in the window is used to replace the pixel at $(x,y)$. 
Mathematically, for an image defined by the function $I(x,y)$ (=pixel intensity at $(x,y)$), a median filter can be defined as follows (square $N \times N$ window):

\begin{equation}
    \label{eqn:median}
    \begin{split}
        I_{filtered}(x,y) & = \text{median}\left(I\left(i,j\right)\right) \text{where} \\ 
        & i \in \left[x-\lfloor N/2 \rfloor,x+\lfloor N/2 \rfloor\right]; \\
        & j \in \left[y-\lfloor N/2 \rfloor,y+\lfloor N/2 \rfloor\right]
    \end{split}
\end{equation}

\subsubsection{Correction Pipeline}
\label{subsubsec:pipeline}
To effectively use all the resources available to us, 
we convert the aforementioned depth correction technique into 
a software pipeline (Figure~\ref{subfig:pipeline}) that uses
multi-threading. We identify three different operations that have 
contrasting requirements. These are as follows: \\
\circled{1} Reading frames from the sensor, mostly an I/O bound operation.\\
\circled{2} Estimating the transform, implemented on the CPU.\\
\circled{3} Combining step, implemented on the GPU.\\
We propose a three-stage pipeline to implement these as separate, overlapping stages to improve the overall efficiency of the system.

\subsection{Epoch Transition}
\label{subsec:adaptive}
After implementing both of our methods, it is necessary to dynamically recalculate the point cloud after the scene has undergone significant changes.
At this point, it is thus necessary to initiate a new {\em epoch}.

We want to perform point cloud fusion as infrequently as possible since it is the slowest component.
The conditional box shown in Figure~\ref{subfig:broadflow} refers to this decision process.
When the scene undergoes significant changes, our previous 
template becomes unsuitable for the current scene, necessitating 
the initiation of a new {\em epoch}. We observed that when 
there is a significant shift
in the scene since the previous fusion 
step, the quality of the matching results deteriorates significantly. 
This phenomenon is studied in Section~\ref{subsubsec:statesswitchfreq}

In our approach, we employ a feature matching technique based on the
ORB method ~\cite{rublee2011orb}. 
ORB uses the 
widely recognized BRIEF ~\cite{calonder2010brief} feature descriptors
in its matching method.
These feature descriptors are vectors that provide a description of a
features in an image, which are then utilized for 
the purpose of matching. 
In order to assess the quality of a match, we calculate the number of 
matches where the {\em feature} distance (Euclidean distance) 
between them exceeds $20$, 
a value that we determined experimentally. 
We call such matches as {\em good matches}.  
The number of good matches between a pair of images is computed and 
if this value is below a predetermined threshold (5 in our experiments),
the system transitions back to the fusion state, initiating 
a new epoch. This process is described in
a timing diagram shown in Figure~\ref{subfig:statediag}.

%% file: evaluation.tex
\section{Results and Analysis}
\label{sec:Evaluation}

\subsection{Setup}
We compare our method with two state-of-the-art research proposals
and present our findings in this section. 
The experimental setup and datasets used in the section have been 
thoroughly described in Section~\ref{sec:Characterization}.

\begin{figure*}[!htb]
\centering
	\includegraphics[width=1.8\columnwidth]{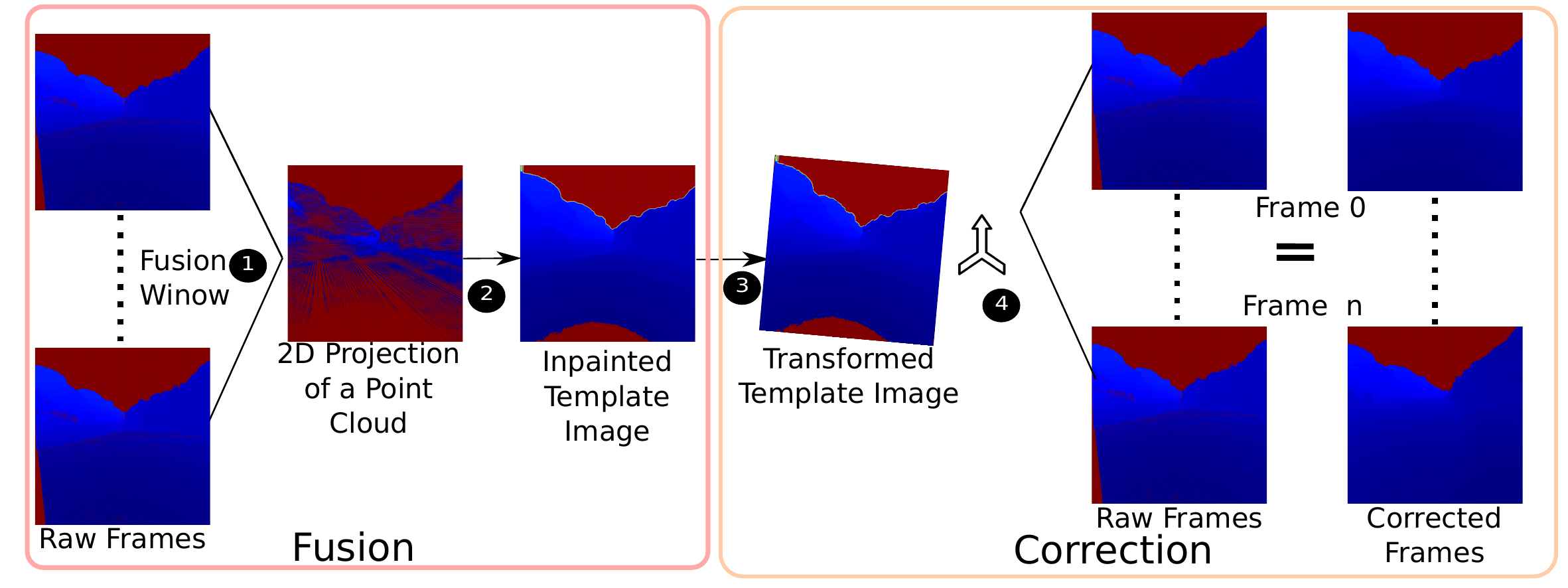}
    \caption{A visual representation of the proposed method with images representing outputs from each of the major steps. (1) First the frames in the fusion window are fused together to form a fused {\em voxel grid}, which is projected back to 2D, (2) The sparse projection is inpainted to create a dense {\em template image}, (3) The template is transformed to align with the current frame and (4) Finally, the incoming raw frame is combined with the template to generate the corrected depth image.}
    \label{fig:flowwithimage}
\end{figure*}

\circled{1} We compare the PSNR and RMSE (root mean squared error) 
with the ground truth depth maps to evaluate the accuracy of the
computed depth maps using different methodologies.
\circled{2} To assess the feasibility of depth correction algorithms on embedded devices,
we analyze their latency and power consumption figures.
\circled{3} In Section~\ref{subsec:sensitivity}, we show two experiments conducted on \voxdepth to 
understand how the size of the fusion window affects image quality and how we developed the necessary conditions
for a state switch.
\circled{4} A simulated 
study is also presented to demonstrate the impact of noise in
depth estimation in a drone swarming task. Figure~\ref{fig:flowwithimage} provides a visual summary of 
\voxdepth including pictorial representations of the outputs of each phase.

For training the state-of-the-art ML method DeepSmooth\cite{krishna2023deepsmooth}, we utilize the Kite Sunny (KTS) dataset. 
It is trained using
the hybrid loss function proposed in the original paper~\cite{krishna2023deepsmooth}.
We utilize the remaining datasets to validate the method and assess its capacity for generalization. We used a gated depth encoder~\cite{yu2019free} and an EfficientNet-lite~\cite{eflite} based 
architecture for the color encoder, as suggested by the authors, to speed up the inference time. Extensive testing was conducted to find the most optimal configuration of 
DeepSmooth that has the least latency and highest quality. We converted the model into the ONNX format~\cite{onnx} and used the CUDA and TensorRT execution providers (in ONNX) 
to achieve 
the best possible latency. 

We also compared our scheme \voxdepth with GSmooth~\cite{islam2018gsmooth} (state-of-the-art non-ML algorithm) and the RealSense~\cite{grunnet2018depth} camera's default hole-filling algorithm. GSmooth detects outliers 
using the least median of squares method in both the spatial and temporal domains. We implemented a CUDA kernel to optimize the latency of GSmooth on the edge device. The 
RealSense SDK includes a hole-filling algorithm -- a simple local image filter that replaces each incorrect pixel with the nearest valid pixel on the left. Every method except DeepSmooth has been implemented with C++ and CUDA.  Deepsmooth used Python and PyTorch. To enable a fair comparison, we only record its inference time.

\subsection{Depth Image Quality}
\label{subsec:quality}

\subsubsection{PSNR}
\label{subsubsec:psnr}

\begin{figure*}[t]
  \centering
\subfloat[Color]{
  \includegraphics[width=0.40\columnwidth]{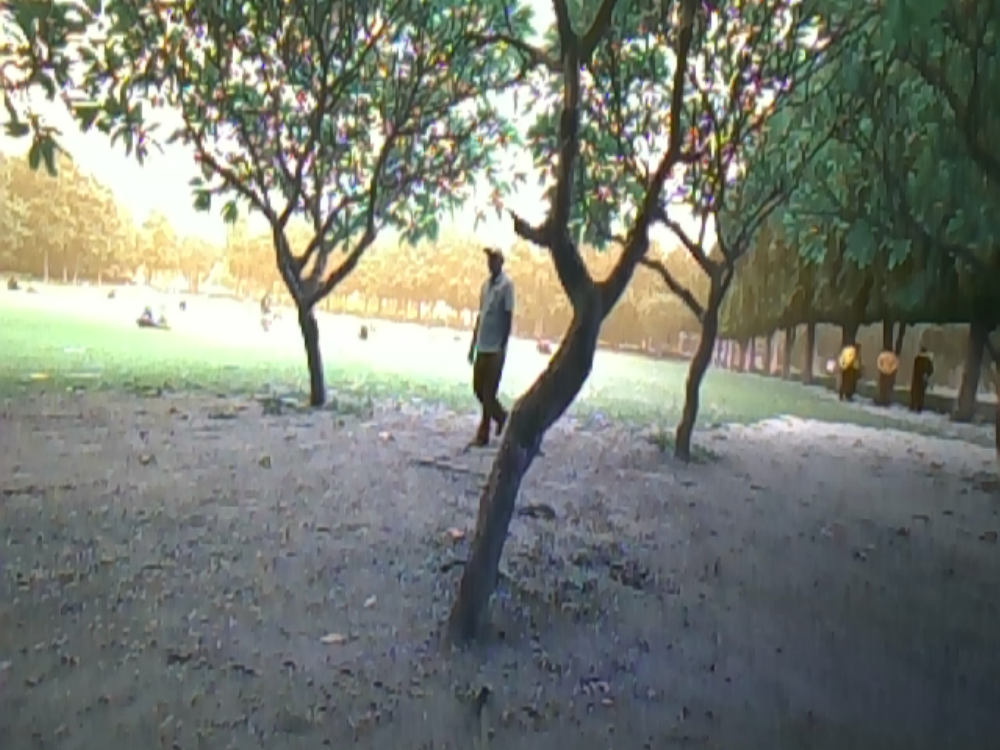}
}
\subfloat[GT]{
  \includegraphics[width=0.40\columnwidth]{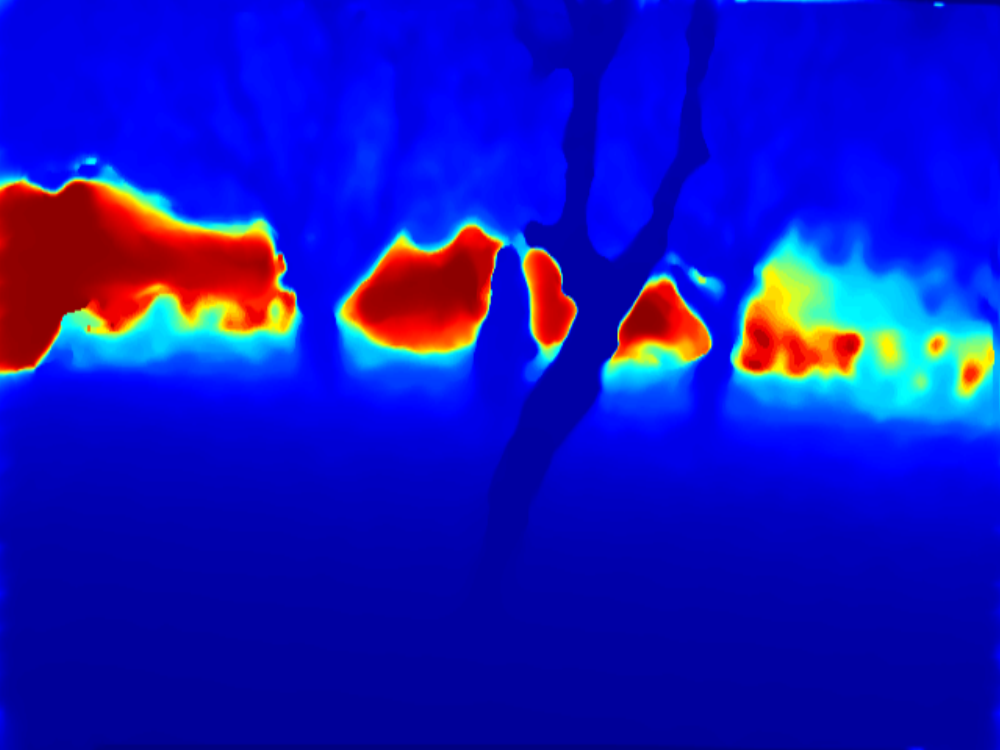}
}
\subfloat[raw]{
  \includegraphics[width=0.40\columnwidth]{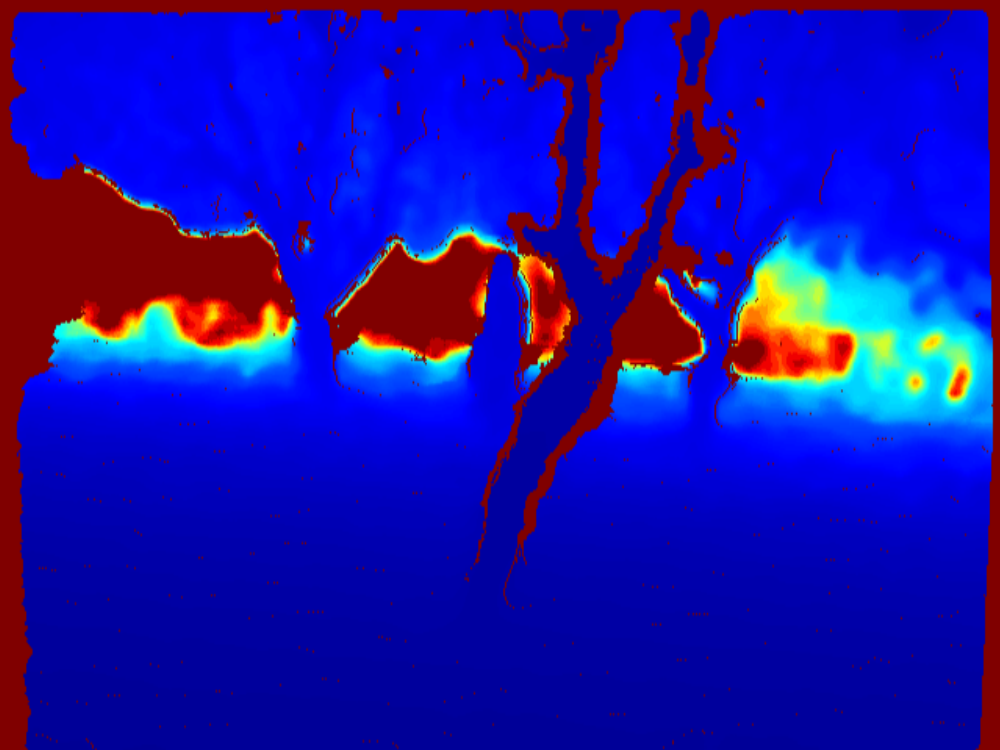}
}
\subfloat[Realsense~\cite{grunnet2018depth}]{
  \includegraphics[width=0.40\columnwidth]{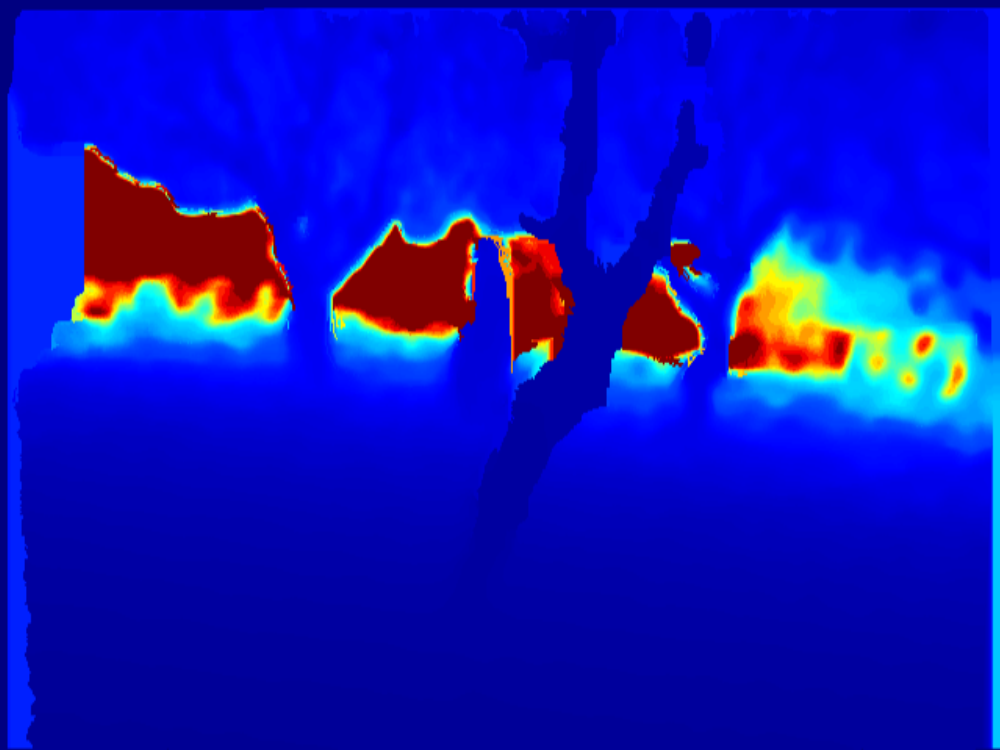}
}
\\
\subfloat[DeepSmooth~\cite{krishna2023deepsmooth}]{
  \includegraphics[width=0.40\columnwidth]{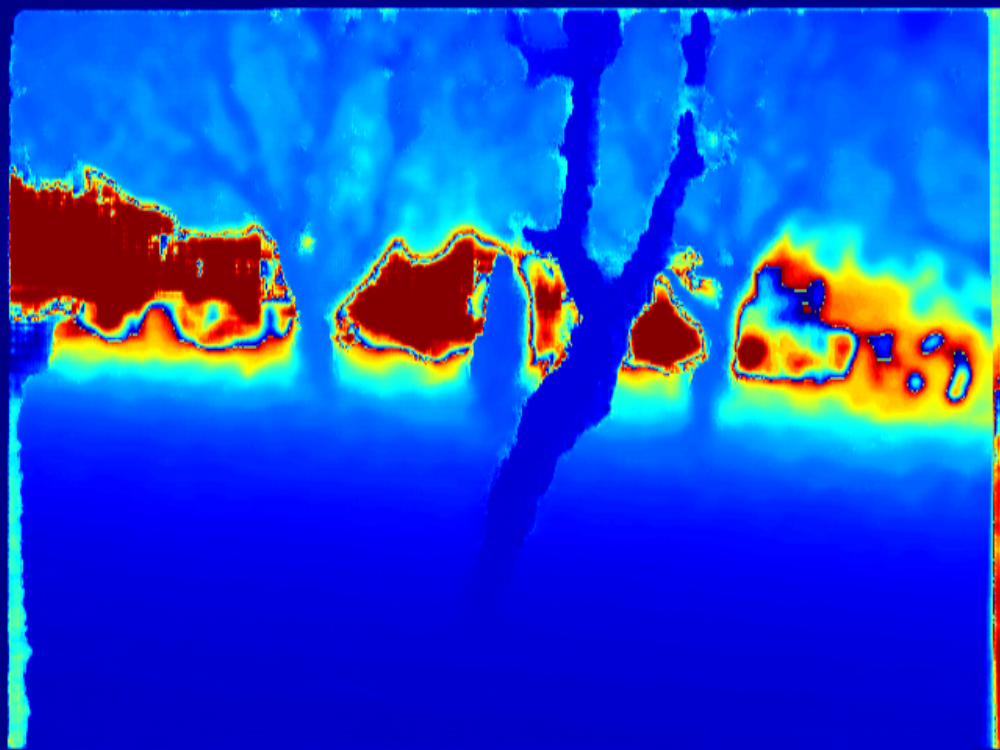}
}
\subfloat[gSmooth~\cite{islam2018gsmooth}]{
  \includegraphics[width=0.40\columnwidth]{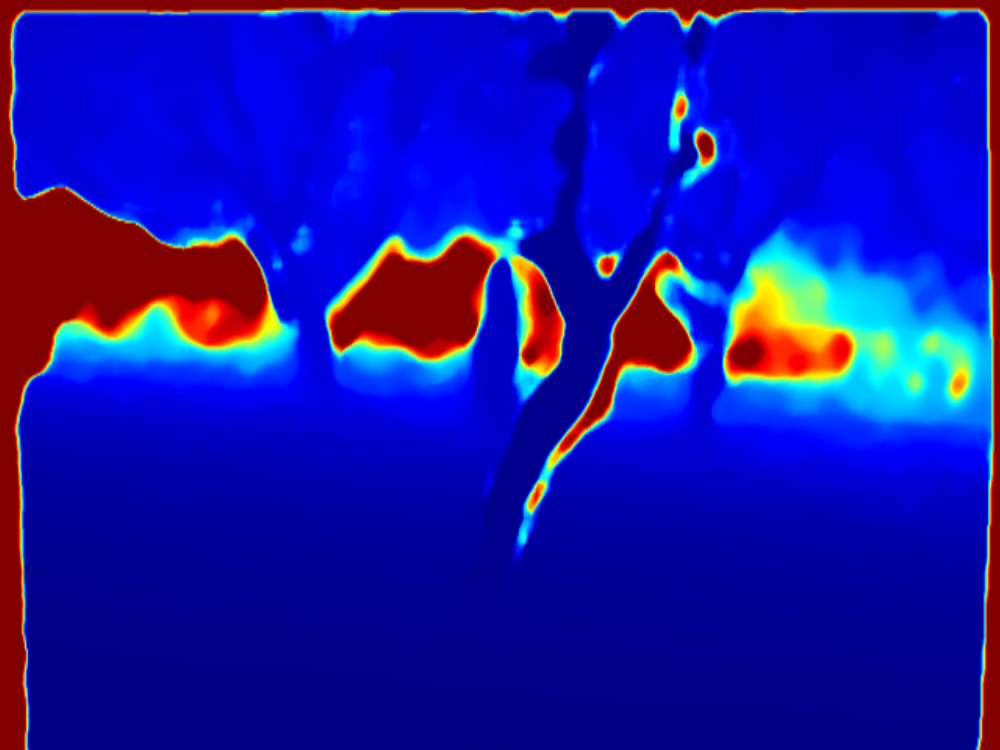}
}
\subfloat[VoxDepth]{
  \includegraphics[width=0.40\columnwidth]{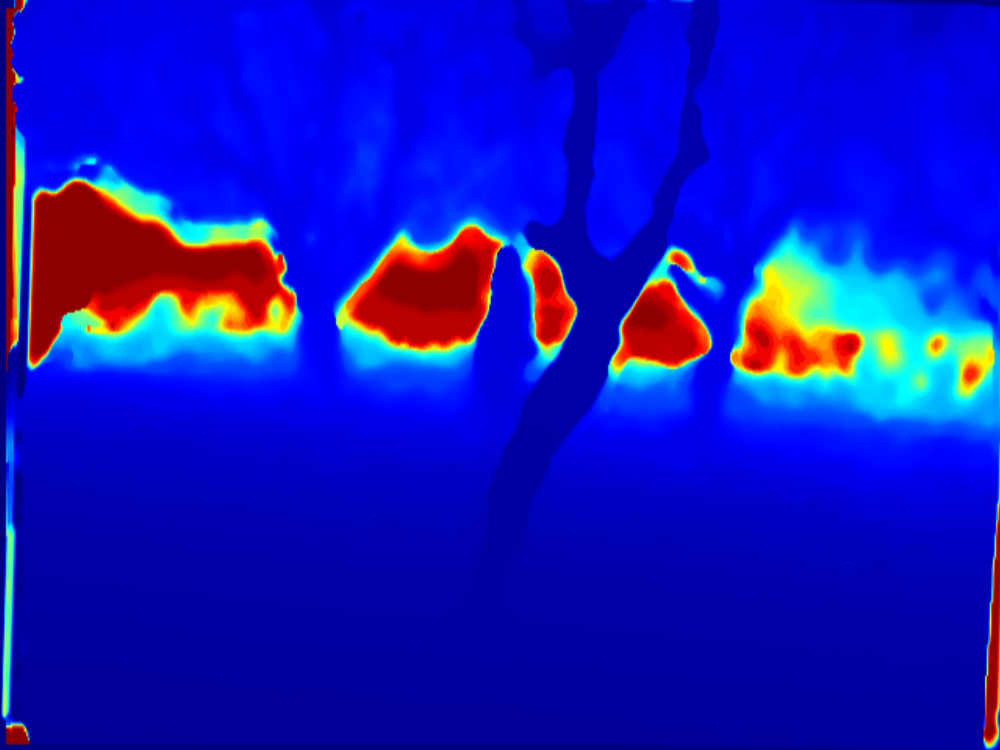}
}
\caption{Performance of different techniques}
\label{fig:compare}
\end{figure*}

The quality of depth maps is measured by calculating the peak signal-to-noise ratio (PSNR) between the generated depth images
and their corresponding ground truth depth images. The PSNR is computed using the mean squared error (MSE) between
corresponding pixels in the ground truth and corrected images. It is
further normalized to the maximum possible pixel value (typically
65,535 for 16-bit depth images). This normalized value is scaled to a logarithmic scale to represent the values in
decibels. Higher PSNR values indicate better image quality. A perfect reconstruction corresponds to an infinite
PSNR.  As presented in Table~\ref{tab:psnr}, \circled{1} in real world datasets, we see a PSNR improvement of $31\%$
over the best competing method (in each dataset) and we also observe
\circled{2} an average improvement of $14.07\%$ on synthetic datasets, except for the Kite
Sunny dataset, where DeepSmooth is better. This shows the superior generalization capabilities of \voxdepth.
\circled{3}
In Figure \ref{fig:compare}, we see that all the methods were able to correct flickering noise in the depth images.

\begin{table}[!htb]
\resizebox{\columnwidth}{!}{%
\begin{tabular}{|l|l|l|l|l|l|}

    \hline
      
    \textbf{Quality} & \textbf{Method} & \textbf{VoxDepth} & \textbf{RealSense} & \textbf{GSmooth} & \textbf{DeepSmooth} \\
     \textbf{Metric} & \textbf{Abbr.} & \textit{VD} & \textit{RS\cite{grunnet2018depth}} & \textit{GS~\cite{islam2018gsmooth}} & \textit{DS~\cite{krishna2023deepsmooth}} \\
    \hline
              & LN & \textbf{17.46} & 13.04 & 13.14 & 14.16  \\
              & MB & \textbf{17.10}  & 13.17 & 15.36 & 10.51 \\
 \textit{PSNR}& KTS & 17.69 & 14.60 & 18.59 & \textbf{22.87} \\
          (dB)    & KTC & \textbf{20.32} & 17.66 & 18.59 & 14.89  \\
              & PLF & \textbf{15.81} & 13.55 & 12.97 & 13.62 \\
              & PLW & \textbf{16.05} & 13.73 & 13.14 & 13.37 \\
    \hline
                & LN & \textbf{112.25} & 167.25 & 179.08 & 180.03 \\
                & MB & 177.61 & \textbf{175.61} & 176.75 & 178.04 \\
\textit{Masked} & KTS & \textbf{102.23} & 139.69 & 172.02 & 175.68 \\
 \textit{RMSE}  & KTC & \textbf{126.15} & 144.70 & 177.93 & 178.89 \\
                & PLF & \textbf{157.95} & 159.15 & 178.70 & 178.38 \\
                & PLW & 169.79 & \textbf{164.79} & 177.59 & 178.87 \\
    \hline
    \end{tabular}

    }
    \caption{Comparison of the quality of different methods based on PSNR (dB, higher is better) and M-RMSE (lower is better). The numbers in bold indicate the best values. \label{tab:psnr}}
\end{table}
    
\subsubsection{Masked RMSE Metric}
\label{subsubsec:masked}
\voxdepth surpasses all other techniques in terms of the accuracy of the estimated depth, 
mostly due to our emphasis on rectifying erroneous patches caused by algorithmic noise
such as occlusion. In order to demonstrate this, we need to accurately measure
the capability of our method to fill the occluded regions with 
the correct depth values. 

By utilizing the masked RMSE metric proposed by Cao et al.~\cite{cao2023scenerf}, we try to evaluate and compare
the depth correction techniques in terms of their capacity to fill the areas that are occluded in one of the cameras.
Specifically, we compute the root mean squared 
error (RMSE) between the pixels of the ground truth depth images and the corrected ones only for
the occluded regions. In Table~\ref{tab:psnr}, we show the results. 
We observe that \circled{1} in most 
benchmarks \voxdepth outperforms the closest competing method (for that dataset)
by an $25.23\%$ on an average, except for the {\em MB} and {\em PLW} benchmarks and \circled{2} even in those 
benchmarks, \voxdepth only lags behind by a small margin of $2.08\%$. 
This supports our claim that \voxdepth has superior occlusion-hole filling capabilities.

\subsection{On-Board Results: Frame Rate and Power Consumption}

\subsubsection{Frame Rate}
\label{subsubsec:framerate}
Another key metric that we need to evaluate
is the frame rate (should be $\ge 20$ FPS).
The results are shown in Table~\ref{tab:fps}.
We observe the following: \circled{1} \voxdepth provides framerates that exceed the requirements for autonomous drones, while also exhibiting strong competitiveness with alternative approaches. A framerate of 26.7 FPS can be provided.
This meets our target.
\circled{2} The pipelined approach exhibits a $21 \%$ speedup as compared to the non-pipelined implementation on the
Jetson Nano device.

\begin{table}[!htb]
  \footnotesize
  \begin{center}
    \begin{tabular}{|l|l|l|l|l|l|l|}
      \hline
     {\textbf{Method}} &{ \textbf{VD}} & {\textbf{VD-}} &  {\textbf{RS~\cite{grunnet2018depth}}} & \textbf{GS ~\cite{islam2018gsmooth}} & \textbf{DS~\cite{krishna2023deepsmooth}} & \textbf{Swarm}    \\
      & & {\textbf{Nopipe}} &  &  &  & \textbf{~\cite{kabore2021distributed}} \\
     \hline

    \textbf{FPS} & 26.71 & 21.91 & \textbf{39.52} & 16.89 & 2 & 14   \\
      \hline
    \end{tabular}
   \end{center}
   \caption{Comparison of the frame rates of different methods on the Jetson Nano}
   \label{tab:fps}
\end{table}

\subsubsection{Component Latency}
\label{subsubsec:componentlatency}
Each of the components described in Section \ref{sec:Implementation} had to adhere to strict time constraints. Otherwise,
it would not have been possible to achieve a framerate of 26.7 FPS. We
list the average latencies for each of these components in Table~\ref{tab:componentlatency}.
The first row shows the numbers for Jetson
Nano. For the purpose of comparison and sanity checking, we
present the results for a workstation PC as well. 
As expected, we observe that the fusion step takes the longest time ($2.9 \times$ slower than the next slowest task that is inpainting).
This can be
explained by the fact that in the fusion step, visual odometry is used to estimate the camera motion in 3D space, and
multiple point clouds are fused together to create a dense scene representation. Inpainting also takes a significant
amount of time because it performs many operations per pixel (maximum in a window). Note that the Fusion and  Inpainting
tasks are performed only once (at the beginning) of an epoch.

\begin{table}[!htb]
  \centering
  \footnotesize
  \caption{Latency of each individual component in the system (in ms). 
  Fusion $\rightarrow$ point cloud fusion, Inpainting $\rightarrow$ depth image inpainting, Transform $\rightarrow$ image registration, Combine $\rightarrow$ combining the template and the frame.}
  \resizebox{1.0\columnwidth}{!}{
  \begin{tabular}{|l|l|l|l|l|}
  \hline
  \textbf{Components} & \textbf{Fusion} & \textbf{Inpainting} & \textbf{Transform} & \textbf{Combine}  \\ 
  \hline
  Jetson Nano & 280 & 94 & 6 & 13 \\ 
  \hline
  Workstation & 85 & 2 & 4 & 3\\ 
  \hline
  \end{tabular}
  }
  \label{tab:componentlatency}
\end{table}

\subsubsection{Power Consumption}
\label{subsubsec:power}

\begin{figure}[!htb]
  \begin{center}
    \includegraphics[width=1.0\columnwidth]{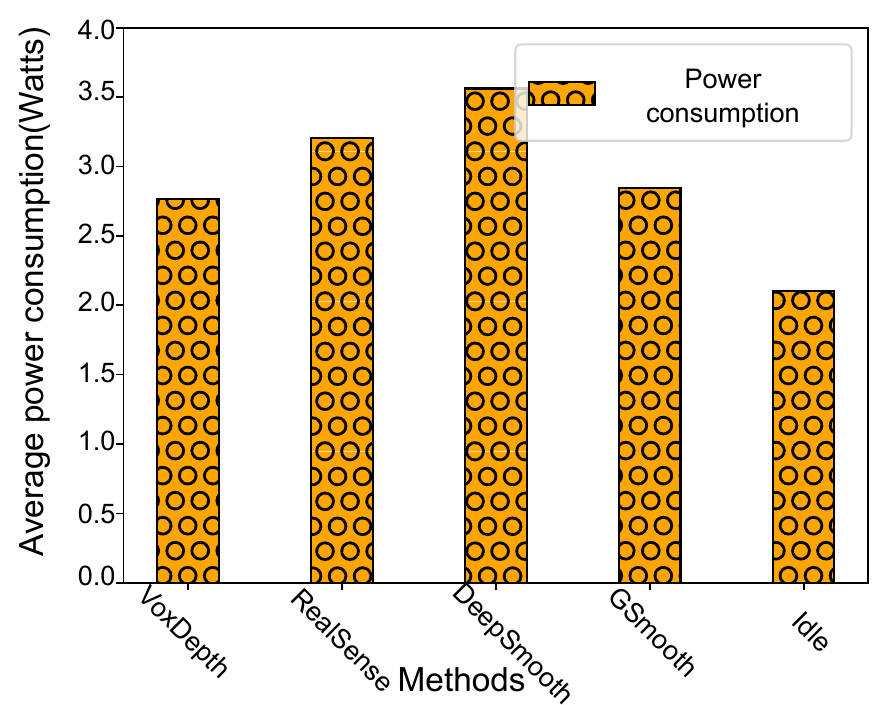}
    \caption{Power consumption of the different techniques \label{fig:power}}
  \end{center}
\end{figure}

We use the {\em jtop}~\cite{Bonghi} tool to get the power usage on the Jetson board.  We calculated the average power
consumption during a specific time frame to assess the power consumption of
a method. Figure~\ref{fig:power} shows a
comparison of various depth correction methods in terms of their average power consumption. 
We find that \circled{1} \voxdepth has $2.9 \%$ lower power consumption than the closest competing
method GSmooth~\cite{islam2018gsmooth} and \circled{2} $28.8 \%$ lower power consumption than the ML-based method
DeepSmooth~\cite{krishna2023deepsmooth}. Because it does not use a heavy neural network for inferencing,
it is more power-efficient than DeepSmooth.  The reason for lower power consumption than GSmooth is because \voxdepth spends far less power in memory operations. Because we use the same 2D template throughout the epoch, we need to store
very little information as compared to competing proposals that rely on much larger stores of information. 

\begin{table*}[!htb]
    \footnotesize
    \begin{center}
        \resizebox{1.0\textwidth}{!}{
            \renewcommand{\arraystretch}{0.8}
            \begin{tabular}{ |l|l|l|l|l|l|l|l|} 
                \hline
                \multirow{2}{*}{\textbf{Year}} &
                \multirow{2}{*}{\textbf{Work}} &
                \textbf{Method} &
                \textbf{ML-} &
                \textbf{Color-} &
                \multirow{2}{*}{\textbf{Point Cloud}} &
                \textbf{Motion} &
				\textbf{    Setup}\\
                
                &
                &
                \textbf{} &
                \textbf{based}&
                \textbf{Guided} &
                &
                \textbf{Comp.} &
				\textbf{@ Frame Rate}
                \\ [0.5ex] 
                \hline\hline
                2011 & Matyunin et al.~\cite{matyunin2011temporal} & Motion Flow Temporal Filter & \textcolor{red}{$\times$}& \textcolor{green}{\checkmark}& \textcolor{red}{$\times$} & \textbf{\textcolor{green}{\checkmark}}  & Intel Celeron 1.8 GHz CPU @ 1.4 fps \\
                \hline
                2016 & Avetisyan et al.~\cite{avetisyan2016temporal} & Optical Flow Temporal Filter & \textcolor{red}{$\times$}& \textcolor{green}{\checkmark}& \textcolor{red}{$\times$}& \textbf{\textcolor{green}{\checkmark}}  & Workstation GPU @ 10 fps\\
                \hline
                2018 & Grunnet-Jepsen et al.~\cite{grunnet2018depth} & Spatial Filter & \textcolor{red}{$\times$}& \textcolor{red}{$\times$}& \textcolor{red}{$\times$}& \textcolor{red}{$\times$}  & Jetson Nano @ 39.52 fps \\
                \hline
                2018 & Islam et al.~\cite{islam2018gsmooth} & Gradient \& LMS-based filter & \textcolor{red}{$\times$}& \textcolor{red}{$\times$}& \textcolor{red}{$\times$} & \textcolor{red}{$\times$} & Jetson Nano @ 16.89 fps \\
                \hline
                2019 & Sterzentsenko et al.~\cite{sterzentsenko2019self} & Multi-sensor Guided Training & \textcolor{green}{\checkmark}& \textcolor{red}{$\times$}&  \textcolor{green}{$\checkmark$}& \textcolor{red}{$\times$} & GeForce GTX 1080 GPU @ 90fps \\
                \hline
                2019 & Chen et al.~\cite{chen2019learning} & 2D-3D feature based & \textcolor{green}{\checkmark}& \textcolor{green}{$\checkmark$}&  \textcolor{green}{$\checkmark$}& \textcolor{red}{$\times$} & \framebox[1.0\width]{Not Available} \\
                \hline
                2021 & Senushkin et al.~\cite{senushkin2021decoder} & Decoder-Modulation CNN& \textcolor{green}{\checkmark}& \textcolor{green}{\checkmark}& \textcolor{red}{$\times$}& \textcolor{red}{$\times$} & Nvidia Tesla P40 GPU  \\
                \hline
                2021 & Imran et al.~\cite{Imran_2021_CVPR} & Surface reconstruction loss& \textcolor{green}{\checkmark}& \textcolor{green}{\checkmark}& \textcolor{red}{$\times$}& \textcolor{red}{$\times$} & GeForce GTX 1080 GPU Ti @ 90fps\\
                \hline
                2023 & Krishna et al.~\cite{krishna2023deepsmooth} & Temporal Encoder & \textcolor{green}{\checkmark}& \textcolor{green}{\checkmark}& \textcolor{red}{$\times$} & \textbf{\textcolor{green}{\checkmark}} &Jetson Nano @ 2fps \\
                \hline
                \textbf{2024} & \textbf{VoxDepth} & 3D Filter  & \textcolor{red}{$\times$} & \textbf{\textcolor{green}{\checkmark}} & \textbf{\textcolor{green}{\checkmark}}& \textbf{\textcolor{green}{\checkmark}}  & Jetson Nano @ 26.71fps \\
                \hline
            \end{tabular}
        }
    \end{center}
    \caption{A comparison of related work}
    \label{table:relwork}
\end{table*}
\label{sec:RelatedWork}

\subsection{Sensitivity Studies}
\label{subsec:sensitivity}

\subsubsection{Fusion Window Size}
\label{subsubsec:winsize}
\begin{figure}[H]
\centering
  \includegraphics[width=0.9\columnwidth]{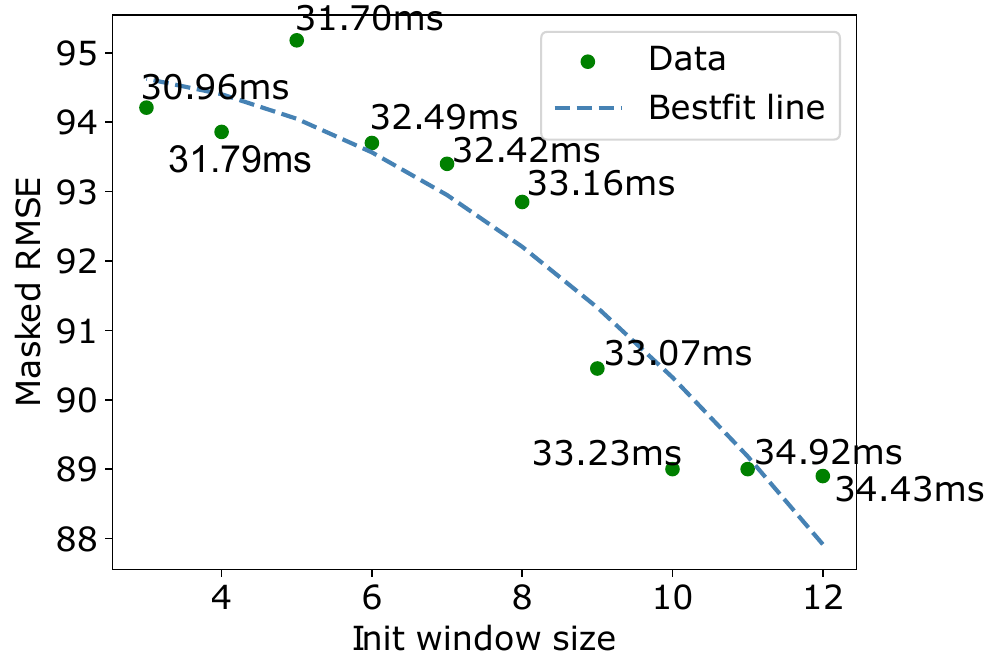}
  \caption{Depth image quality vs voxel fusion
window size. The number against each data point shows the
average time taken to generate a single frame in ms.}
  \label{fig:winsize}
\end{figure}

To understand what the ideal number of frames should be in the fusion window,
we plot the quality of depth images for different fusion window sizes in Figure \ref{fig:winsize}. We also annotate the average time taken to generate a single frame beside
each data point. We find that there is a clear trade-off between quality and latency. In our experiments, we use a 
frame window of
size 10 for voxel fusion. 
This is the most optimal configuration. 
\circled{1} A $10$ frame fusion window, when compared to the largest window size of $12$, generates similar quality 
depth images but reduces latency by $3.61\%$.

\subsubsection{State Switch: Optimal Time and Frequency}
\label{subsubsec:statesswitchfreq}
We measured the average pixel movement between frames by generating an optical flow (refer to Section~\ref{subsec:opticalflow}).
We found that RGB images 50 frames
apart showed an average pixel movement of $6.75$ pixels and the quality of registration (see 
Section~\ref{subsec:resize}) was
$16.06$ dB. When we repeated this experiment with images 150 frames apart,
the average pixel movement went up to $10.08$ pixels, and the quality of registration went
down to $12.61$ decibels.
The conclusion here is that there is a strong relationship between these three variables (positive or negative):
the distance between frames, the average pixel movement and the image registration quality. 

The frequency of switching between fusion and correction states (shown in Figure \ref{subfig:statediag}) depends on the
dataset in use, the type of motion in the dataset and the amount of textural information in the scene.  With all other
parameters constant, the number of state switches recorded for each dataset is presented in Table
\ref{tab:switchfrequency}.  In most cases, the synthetically generated datasets require fewer switches due to consistent
lighting and contrast values across frames. Except the Kite Sunny dataset, we observed a $40.33\%$ lower number of state
switches in synthetically generated datasets as compared to datasets generated using the RealSense cameara.  One reason
behind high frequency of switching would be fast changes in the scene that the camera sees. Such a scenario is unlikely
to happen in the case of autonomous drones.

The insights in this study led us to the algorithm that decides when to end an epoch and start a new one (state switch). 

\begin{table}[!htb]
 \footnotesize
  \centering
  \caption{Number of state switches for each dataset while processing 500 frames}
  \begin{tabular}{|l|l|l|l|l|l|}
  \hline
  \textbf{LN} & \textbf{MB} & \textbf{KTS} & \textbf{KTC} & \textbf{PLF}& \textbf{PLW} \\ 
  \hline
  10 & 10 & 4 & 14 & 4 & 5\\ 
  \hline
  \end{tabular}
  \label{tab:switchfrequency}
\end{table}

\subsubsection{Impact of the  Depth Estimation Noise on Drone Swarming}
\label{subsec:consequences}

In order to strengthen our thesis about the importance of having
precise depth images, we introduced distorted depth data into a 
drone swarming simulator (SmrtSwarm~\cite{bhamu2023smrtswarm}). 
We used a city skyline scene with a leader drone and seven drones following the leader in a swarming formation.
Each drone is equipped with a depth estimation device (simulated stereo camera) to measure its distance from obstacles and other drones.
The collision ratio is defined as the proportion of simulated runs that result in collisions divided
by the total number of runs. 
We conducted these simulated runs by introducing varying levels of sensor noise 
to the simulated depth measurement camera. 
This noise is defined by a noise ratio threshold, $\theta$. A noise ratio, $\beta$ is uniformly sampled from the range $\left[-\theta,+\theta\right]$. The noise ratio is subsequently multiplied by the depth measurement and added to the depth in order to obtain the new depth value. $\theta$ in our experiments is a single precision floating point number varying from $0.3$ to $1.0$. 
In Equation~\ref{eqn:noisesim}, $D_{true}$ is the depth measurement value from the simulated stereo camera.
$D_{noise}$ is the noise added depth measurement.

\begin{equation}
    D_{noise} = D_{true} + D_{true} \times \beta
    \label{eqn:noisesim}
\end{equation}

\begin{figure}[!htb]
  \centering
  \includegraphics[width=1.0\columnwidth]{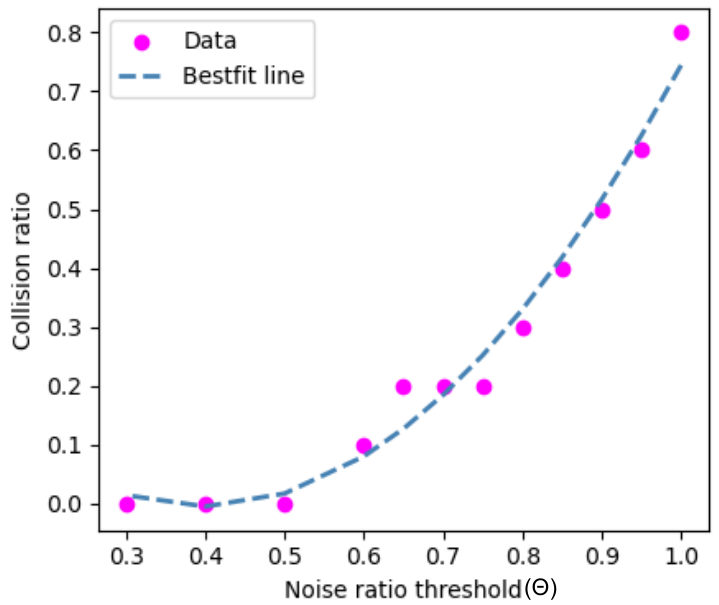}
  \caption{Relation between the number of collisions in simulations and the depth measurement noise}
  \label{fig:collide}
\end{figure}

We present our findings in Figure~\ref{fig:collide}. 
\circled{1} As noise increases, the collision frequency grows super-linearly (almost quadratically).
\circled{2} After a $\theta$ exceeds $0.5$, increasing $\theta$ by $2\times$ leads to an $8\times$ increase in  the number of collisions.

%% file: relatedwork.tex
\section{Related Work}
\label{sec:relwork}

Over the past few years, there have been several notable attempts to tackle this problem by exploiting spatial and temporal similarities in depth videos. 
Ibrahim et al~\cite{ibrahim2020depth} present a comprehensive list of many such methods in their survey paper.
We also present a brief comparison of the most relevant related work in Table ~\ref{table:relwork}.
Previous work in this field can be classified based on the following features: utilization of point clouds, motion
compensation, the usage of RGB channels in depth correction and the type of the technique (ML or non-ML).  

\subsection{ML Methods}
\label{sec:mlmethod}
ML models in this space are associated with a lot of issues.
There is a paucity of realworld datasets and the algorithms are often quite slow and power consuming.
They are thus not suitable for edge devices like the Jetson Nano board.
Sterzentsenko et al.~\cite{sterzentsenko2019self} show that the issue of lack
of ground truth data can be tackled by having multiple sensors capture
the same scene (multi-sensor fusion). This allows the system to create its own semi-synthetic dataset
by first creating a highly accurate 3D representation of the scene by fusing the information from multiple viewpoints,
and then generating a series of 2D or 3D scenes. 
This method is effective in small enclosed spaces, but it
cannot be implemented in open outdoor spaces because of the large number of
sensors required (each of which would cost over $\$400$ USD) and the other environmental 
limitations traditionally associated with an outdoor
environment.

Most ML approaches ~\cite{krishna2023deepsmooth,zhang2018deep,xu2019depth} use an encoder-decoder CNN architecture similar to Unet~\cite{ronneberger2015u}. 
The encoder is responsible for feature extraction whereas the decoder uses those features to improve the original depth
map. Some works like DeepSmooth~\cite{krishna2023deepsmooth}, also suggest the use of two encoder branches: one for 
color and the other for depth images to train the network. While this technique produces highly accurate depth images, we
have shown in Section~\ref{subsubsec:framerate} that this method is slow (least performing in our evaluation).
The framerate is way lower than 20 FPS (our minimum threshold). 

\subsection{Non-ML Methods}
\label{sec:nonmlmethods}
Non-ML methods tend to generalize better than their ML counterparts, which are heavily
dependent on the amount, diversity and quality of the baseline
datasets. 
The bulk of the work in this domain relies on classical
filters. The generic idea is to detect and flag erroneous pixels 
by identifying atypical features/neighborhoods (both temporal and spatial).
These flagged pixels are then replaced by predicted depth values. 

Grunnet et al.~\cite{grunnet2018depth} only use the
spatial neighborhood of the depth image to fix holes in the frame (a spatial filter).
This proves to be the least computationally expensive 
approach, and hence ships with the Intel Realsense SDK. But these methods fail to perform as we have
observed in 
Section~\ref{subsec:quality}.

Few approaches in this domain~\cite{matyunin2011temporal,avetisyan2016temporal}, 
use previous frames in the depth videos to both detect and replace erroneous pixels. This is  a temporal filter.
To account for motion within the 
frame as well camera movement, they also suggest compensating the motion with methods such as optical flow.
This results in a considerable amount of latency on 
smaller devices.
Islam et al.~\cite{islam2018gsmooth} combine the spatial and temporal filters in their work to filter out erroneous pixels. They do not, however, suggest any motion 
compensation in favor of faster processing. This results in a considerable hit in terms of
quality when compared to \voxdepth (refer to Section~\ref{subsec:quality}).

%% file: conclusion.tex
\section{Conclusion}
\label{sec:Conclusion}
The key premise of our paper was that existing ML methods are accurate yet slow, whereas non-ML techniques are quite
fast but do not provide adequate quality. Given that many works in the edge computing domain have pointed out that a
good frame rate (at least 20 FPS) is required, there was a strong need to develop such a solution that can run on
embedded boards.  Our proposal, \voxdepth, was able to successfully provide a frame rate of 27 FPS on an NVIDIA Jetson
Nano board and also outperform the state of the art (both ML-based and non-ML) in terms of the depth image quality.  It
particularly did very well in removing algorithmic noise (improvement in masked RMSE by 25\%) and PSNR (31\% better). It
also proved to be 58\% faster that the closest competing proposal in the literature. 

The key learnings from the paper are as follows:

\begin{enumerate}
\item To rectify 2D depth images, it is a good idea to maintain a 3D representation of the scene in the form of 
a point cloud. It preserves important information. 

\item Instead of relying on a lot of ephemeral data, it is a better idea to split time into epochs and use a single
fused point cloud as the basis for scene rectification throughout an epoch. It provides a stable baseline. 

\item Converting the fused point cloud to a template image has two key advantages. First it allows us to use standard
image registration techniques, and second, it is very performance efficient (faster than 3D$\leftrightarrow$2D comparison). 

\item Algorithmic holes are an important source of noise and are fundamentally different from random flickering noise. They
represent the systematic component of noise. This work rightly takes cognizance of them and also uses metrics like the masked
RMSE metric to specifically assess whether they have been properly filled or not. 

\item Using a pipeline approach is a wise idea in a heterogeneous system that comprises CPUs, GPUs and accelerators. It ensures that all the parts of the system are used and there is no idling. 
\end{enumerate}

%% file: main.bbl
\begin{thebibliography}{10}
\providecommand{\url}[1]{#1}
\csname url@samestyle\endcsname
\providecommand{\newblock}{\relax}
\providecommand{\bibinfo}[2]{#2}
\providecommand{\BIBentrySTDinterwordspacing}{\spaceskip=0pt\relax}
\providecommand{\BIBentryALTinterwordstretchfactor}{4}
\providecommand{\BIBentryALTinterwordspacing}{\spaceskip=\fontdimen2\font plus
\BIBentryALTinterwordstretchfactor\fontdimen3\font minus \fontdimen4\font\relax}
\providecommand{\BIBforeignlanguage}[2]{{%
\expandafter\ifx\csname l@#1\endcsname\relax
\typeout{** WARNING: IEEEtranS.bst: No hyphenation pattern has been}%
\typeout{** loaded for the language `#1'. Using the pattern for}%
\typeout{** the default language instead.}%
\else
\language=\csname l@#1\endcsname
\fi
#2}}
\providecommand{\BIBdecl}{\relax}
\BIBdecl

\bibitem{jetsonagxorin}
\BIBentryALTinterwordspacing
[Online; accessed 2024-06-11]. [Online]. Available: \url{https://www.nvidia.com/en-in/autonomous-machines/embedded-systems/jetson-orin/}
\BIBentrySTDinterwordspacing

\bibitem{eflite}
\BIBentryALTinterwordspacing
``Higher accuracy on vision models with efficientnet-lite,'' [Online; accessed 2024-03-30]. [Online]. Available: \url{https://blog.tensorflow.org/2020/03/higher-accuracy-on-vision-models-with-efficientnet-lite.html}
\BIBentrySTDinterwordspacing

\bibitem{onnx}
\BIBentryALTinterwordspacing
``Onnx | home,'' [Online; accessed 2024-06-11]. [Online]. Available: \url{https://onnx.ai/}
\BIBentrySTDinterwordspacing

\bibitem{TegraLinuxDriver}
\BIBentryALTinterwordspacing
``Tegra linux driver,'' [Online; accessed 2024-03-30]. [Online]. Available: \url{https://docs.nvidia.com/jetson/archives/l4t-archived/l4t-325/index.html}
\BIBentrySTDinterwordspacing

\bibitem{assunccao2022real}
E.~Assun{\c{c}}{\~a}o, P.~D. Gaspar, R.~Mesquita, M.~P. Sim{\~o}es, K.~Alibabaei, A.~Veiros, and H.~Proen{\c{c}}a, ``Real-time weed control application using a jetson nano edge device and a spray mechanism,'' \emph{Remote Sensing}, vol.~14, no.~17, p. 4217, 2022.

\bibitem{avetisyan2016temporal}
R.~Avetisyan, C.~Rosenke, M.~Luboschik, and O.~Staadt, ``Temporal filtering of depth images using optical flow,'' 2016.

\bibitem{bay2006surf}
H.~Bay, T.~Tuytelaars, and L.~Van~Gool, ``Surf: Speeded up robust features,'' in \emph{Computer Vision--ECCV 2006: 9th European Conference on Computer Vision, Graz, Austria, May 7-13, 2006. Proceedings, Part I 9}.\hskip 1em plus 0.5em minus 0.4em\relax Springer, 2006, pp. 404--417.

\bibitem{bhamu2023smrtswarm}
N.~Bhamu, H.~Verma, A.~Dixit, B.~Bollard, and S.~R. Sarangi, ``Smrtswarm: A novel swarming model for real-world environments,'' \emph{Drones}, vol.~7, no.~9, p. 573, 2023.

\bibitem{Bonghi}
\BIBentryALTinterwordspacing
R.~Bonghi, ``Jetson-stats,'' [Online; accessed 2024-03-30]. [Online]. Available: \url{https://rnext.it/jetson_stats/}
\BIBentrySTDinterwordspacing

\bibitem{buades2005non}
A.~Buades, B.~Coll, and J.-M. Morel, ``A non-local algorithm for image denoising,'' in \emph{2005 IEEE CVPR}, vol.~2.\hskip 1em plus 0.5em minus 0.4em\relax Ieee, 2005, pp. 60--65.

\bibitem{calonder2010brief}
M.~Calonder, V.~Lepetit, C.~Strecha, and P.~Fua, ``Brief: Binary robust independent elementary features,'' in \emph{Computer Vision--ECCV 2010}.\hskip 1em plus 0.5em minus 0.4em\relax Springer, 2010, pp. 778--792.

\bibitem{cao2023scenerf}
A.-Q. Cao and R.~de~Charette, ``Scenerf: Self-supervised monocular 3d scene reconstruction with radiance fields,'' in \emph{Proceedings of CVPR}, 2023, pp. 9387--9398.

\bibitem{chen2019learning}
Y.~Chen, B.~Yang, M.~Liang, and R.~Urtasun, ``Learning joint 2d-3d representations for depth completion,'' in \emph{Proceedings of CVPR}, 2019, pp. 10\,023--10\,032.

\bibitem{dai2020rgb}
W.~Dai, Y.~Zhang, P.~Li, Z.~Fang, and S.~Scherer, ``Rgb-d slam in dynamic environments using point correlations,'' \emph{IEEE Transactions on Pattern Analysis and Machine Intelligence}, vol.~44, no.~1, pp. 373--389, 2020.

\bibitem{delmerico2019we}
J.~Delmerico, T.~Cieslewski, H.~Rebecq, M.~Faessler, and D.~Scaramuzza, ``Are we ready for autonomous drone racing? the uzh-fpv drone racing dataset,'' in \emph{2019 International Conference on Robotics and Automation (ICRA)}.\hskip 1em plus 0.5em minus 0.4em\relax IEEE, 2019, pp. 6713--6719.

\bibitem{ferstl2013image}
D.~Ferstl, C.~Reinbacher, R.~Ranftl, M.~R{\"u}ther, and H.~Bischof, ``Image guided depth upsampling using anisotropic total generalized variation,'' in \emph{Proceedings of the IEEE international conference on computer vision}, 2013, pp. 993--1000.

\bibitem{fonder2019mid}
M.~Fonder and M.~Van~Droogenbroeck, ``Mid-air: A multi-modal dataset for extremely low altitude drone flights,'' in \emph{Proceedings of CVPR workshops}, 2019, pp. 0--0.

\bibitem{grunnet2018depth}
A.~Grunnet-Jepsen and D.~Tong, ``Depth post-processing for intel{\textregistered} realsense™ d400 depth cameras,'' \emph{New Technologies Group, Intel Corporation}, vol.~3, 2018.

\bibitem{ibrahim2020depth}
M.~M. Ibrahim, Q.~Liu, R.~Khan, J.~Yang, E.~Adeli, and Y.~Yang, ``Depth map artefacts reduction: A review,'' \emph{IET Image Processing}, vol.~14, no.~12, pp. 2630--2644, 2020.

\bibitem{Imran_2021_CVPR}
S.~Imran, X.~Liu, and D.~Morris, ``Depth completion with twin surface extrapolation at occlusion boundaries,'' in \emph{Proceedings of CVPR (CVPR)}, June 2021, pp. 2583--2592.

\bibitem{islam2018gsmooth}
A.~T. Islam, M.~Luboschik, A.~Jirka, and O.~Staadt, ``gsmooth: A gradient based spatial and temporal method of depth image enhancement,'' in \emph{Proceedings of Computer Graphics International 2018}, 2018, pp. 175--184.

\bibitem{kabore2021distributed}
K.~M. Kabore and S.~G{\"u}ler, ``Distributed formation control of drones with onboard perception,'' \emph{IEEE/ASME Transactions on Mechatronics}, vol.~27, no.~5, pp. 3121--3131, 2021.

\bibitem{keselman2017intel}
L.~Keselman, J.~Iselin~Woodfill, A.~Grunnet-Jepsen, and A.~Bhowmik, ``Intel realsense stereoscopic depth cameras,'' in \emph{Proceedings of the IEEE CVPR workshops}, 2017, pp. 1--10.

\bibitem{kim2021study}
J.-H. Kim, T.-H. Lee, Y.~Han, and H.~Byun, ``A study on the design and implementation of multi-disaster drone system using deep learning-based object recognition and optimal path planning,'' \emph{KIPS Transactions on Computer and Communication Systems}, vol.~10, no.~4, pp. 117--122, 2021.

\bibitem{kopf2007joint}
J.~Kopf, M.~F. Cohen, D.~Lischinski, and M.~Uyttendaele, ``Joint bilateral upsampling,'' \emph{ACM Transactions on Graphics (ToG)}, vol.~26, no.~3, pp. 96--es, 2007.

\bibitem{krishna2023deepsmooth}
S.~Krishna and B.~S. Vandrotti, ``Deepsmooth: Efficient and smooth depth completion,'' in \emph{Proceedings of CVPR}, 2023, pp. 3357--3366.

\bibitem{lane1994stereo}
R.~Lane and N.~Thacker, ``Stereo vision research: An algorithm survey,'' \emph{URL citeseer. ist. psu. edu/lane96stereo. html}, 1994.

\bibitem{lasota2014toward}
P.~A. Lasota, G.~F. Rossano, and J.~A. Shah, ``Toward safe close-proximity human-robot interaction with standard industrial robots,'' in \emph{2014 IEEE CASE}.\hskip 1em plus 0.5em minus 0.4em\relax IEEE, 2014, pp. 339--344.

\bibitem{luebke2008cuda}
D.~Luebke, ``Cuda: Scalable parallel programming for high-performance scientific computing,'' in \emph{2008 5th IEEE international symposium on biomedical imaging: from nano to macro}.\hskip 1em plus 0.5em minus 0.4em\relax IEEE, 2008, pp. 836--838.

\bibitem{matyunin2011temporal}
S.~Matyunin, D.~Vatolin, Y.~Berdnikov, and M.~Smirnov, ``Temporal filtering for depth maps generated by kinect depth camera,'' in \emph{2011 3DTV Conference: The True Vision-Capture, Transmission and Display of 3D Video (3DTV-CON)}.\hskip 1em plus 0.5em minus 0.4em\relax IEEE, 2011, pp. 1--4.

\bibitem{mizukami2007optical}
Y.~Mizukami and K.~Tadamura, ``Optical flow computation on compute unified device architecture,'' in \emph{14th International Conference on Image Analysis and Processing (ICIAP 2007)}.\hskip 1em plus 0.5em minus 0.4em\relax IEEE, 2007, pp. 179--184.

\bibitem{amrm}
MordorIntelligence, ``Autonomous mobile robots market,'' \url{https://www.mordorintelligence.com/industry-reports/autonomous-mobile-robot-market}, [Online; accessed 2024-01-20].

\bibitem{muja2009flann}
M.~Muja and D.~Lowe, ``Flann-fast library for approximate nearest neighbors user manual,'' \emph{Computer Science Department, University of British Columbia, Vancouver, BC, Canada}, vol.~5, no.~6, 2009.

\bibitem{ronneberger2015u}
O.~Ronneberger, P.~Fischer, and T.~Brox, ``U-net: Convolutional networks for biomedical image segmentation,'' in \emph{Medical Image Computing and Computer-Assisted Intervention--MICCAI 2015}.\hskip 1em plus 0.5em minus 0.4em\relax Springer, 2015, pp. 234--241.

\bibitem{rublee2011orb}
E.~Rublee, V.~Rabaud, K.~Konolige, and G.~Bradski, ``Orb: An efficient alternative to sift or surf,'' in \emph{2011 International conference on computer vision}.\hskip 1em plus 0.5em minus 0.4em\relax Ieee, 2011, pp. 2564--2571.

\bibitem{sa2014improved}
Y.~Sa, ``Improved bilinear interpolation method for image fast processing,'' in \emph{2014 7th International Conference on Intelligent Computation Technology and Automation}.\hskip 1em plus 0.5em minus 0.4em\relax IEEE, 2014, pp. 308--311.

\bibitem{senushkin2021decoder}
D.~Senushkin, M.~Romanov, I.~Belikov, N.~Patakin, and A.~Konushin, ``Decoder modulation for indoor depth completion,'' in \emph{2021 IEEE/RSJ IROS}.\hskip 1em plus 0.5em minus 0.4em\relax IEEE, 2021, pp. 2181--2188.

\bibitem{shah2018airsim}
S.~Shah, D.~Dey, C.~Lovett, and A.~Kapoor, ``Airsim: High-fidelity visual and physical simulation for autonomous vehicles,'' in \emph{Field and Service Robotics: Results of the 11th International Conference}.\hskip 1em plus 0.5em minus 0.4em\relax Springer, 2018, pp. 621--635.

\bibitem{song2014sliding}
S.~Song and J.~Xiao, ``Sliding shapes for 3d object detection in depth images,'' in \emph{Computer Vision--ECCV 2014: 13th European Conference}.\hskip 1em plus 0.5em minus 0.4em\relax Springer, 2014, pp. 634--651.

\bibitem{steinbrucker2011real}
F.~Steinbr{\"u}cker, J.~Sturm, and D.~Cremers, ``Real-time visual odometry from dense rgb-d images,'' in \emph{2011 IEEE ICCV Workshops}.\hskip 1em plus 0.5em minus 0.4em\relax IEEE, 2011, pp. 719--722.

\bibitem{sterzentsenko2019self}
V.~Sterzentsenko, L.~Saroglou, A.~Chatzitofis, S.~Thermos, N.~Zioulis, A.~Doumanoglou, D.~Zarpalas, and P.~Daras, ``Self-supervised deep depth denoising,'' in \emph{Proceedings of CVPR}, 2019, pp. 1242--1251.

\bibitem{teed2020raft}
Z.~Teed and J.~Deng, ``Raft: Recurrent all-pairs field transforms for optical flow,'' in \emph{Computer Vision--ECCV 2020: 16th European Conference, Glasgow, UK, August 23--28, 2020, Proceedings, Part II 16}.\hskip 1em plus 0.5em minus 0.4em\relax Springer, 2020, pp. 402--419.

\bibitem{Thadani_Siddiqui_Lerman_Merrill_2023}
\BIBentryALTinterwordspacing
T.~Thadani, F.~Siddiqui, R.~Lerman, and J.~B. Merrill, ``Tesla drivers run autopilot where it’s not intended — with deadly consequences,'' Dec 2023, [Online; accessed 2024-03-30]. [Online]. Available: \url{https://www.washingtonpost.com/technology/2023/12/10/tesla-autopilot-crash/}
\BIBentrySTDinterwordspacing

\bibitem{tzimiropoulos2010robust}
G.~Tzimiropoulos, V.~Argyriou, S.~Zafeiriou, and T.~Stathaki, ``Robust fft-based scale-invariant image registration with image gradients,'' \emph{IEEE transactions on pattern analysis and machine intelligence}, vol.~32, no.~10, pp. 1899--1906, 2010.

\bibitem{Uraizee_2023}
\BIBentryALTinterwordspacing
Uraizee, T.~T.~R. Lerman, I.~Piper, F.~Siddiqui, and Irfan, ``Inside the final seconds of a deadly tesla autopilot crash,'' Oct 2023, [Online; accessed 2024-03-30]. [Online]. Available: \url{https://www.washingtonpost.com/technology/interactive/2023/tesla-autopilot-crash-analysis/}
\BIBentrySTDinterwordspacing

\bibitem{vincent1993morphological}
L.~Vincent, ``Morphological grayscale reconstruction in image analysis: applications and efficient algorithms,'' \emph{IEEE transactions on image processing}, vol.~2, no.~2, pp. 176--201, 1993.

\bibitem{viswanathan2009features}
D.~G. Viswanathan, ``Features from accelerated segment test (fast),'' in \emph{Proceedings of the 10th workshop on image analysis for multimedia interactive services, London, UK}, 2009, pp. 6--8.

\bibitem{xu2019depth}
Y.~Xu, X.~Zhu, J.~Shi, G.~Zhang, H.~Bao, and H.~Li, ``Depth completion from sparse lidar data with depth-normal constraints,'' in \emph{Proceedings of CVPR}, 2019, pp. 2811--2820.

\bibitem{yu2019free}
J.~Yu, Z.~Lin, J.~Yang, X.~Shen, X.~Lu, and T.~S. Huang, ``Free-form image inpainting with gated convolution,'' in \emph{Proceedings of the IEEE/CVF international conference on computer vision}, 2019, pp. 4471--4480.

\bibitem{zhang2018deep}
Y.~Zhang and T.~Funkhouser, ``Deep depth completion of a single rgb-d image,'' in \emph{Proceedings of CVPR}, 2018, pp. 175--185.

\end{thebibliography}
